\documentclass[11pt]{article}
\usepackage[margin=1in]{geometry} 
\usepackage{geometry}
\usepackage{layout} 
\usepackage{float}
\usepackage{fancyhdr}
\usepackage{enumitem}
\usepackage{booktabs}
\usepackage{latexsym}
\usepackage{array}
\newcommand{\citereserveBackslash}[1]{\let\temp=\\#1\let\\=\temp}
\newcolumntype{C}[1]{>{\citereserveBackslash\centering}p{#1}}
\newcolumntype{R}[1]{>{\citereserveBackslash\raggedleft}p{#1}}
\newcolumntype{L}[1]{>{\citereserveBackslash\raggedright}p{#1}}
\usepackage{longtable}
\usepackage{appendix}
\usepackage{supertabular}
\usepackage{multirow}
\usepackage{graphicx}
\usepackage{caption}
\usepackage{url}
\usepackage{hyperref}
\usepackage{subfigure}
\usepackage{amsmath}
\usepackage{bbm}
\allowdisplaybreaks[4]
\usepackage{amssymb}
\usepackage{listings}
\usepackage{titlesec}
\usepackage[usestackEOL]{stackengine}
\titlespacing*{\citearagraph} {15pt}{6pt}{1em}
\titlespacing*{\subparagraph} {15pt}{6pt}{1em}
\titlespacing*{\subsection} {5pt}{8pt}{3pt}
\titlespacing*{\subsubsection} {8pt}{8pt}{3pt}

\newcommand{\wordscore}[2]{%
    \stackunder[2pt]{%
        \textbf{\textcolor{red}{#1}}
    }{%
        \scriptsize \textcolor{gray}{(#2)}
    }%
}

\usepackage[
  backend=biber,
  natbib,
  citestyle=authoryear,
  style=apa,
  uniquelist=false,
  uniquename=false,
  sorting=nyt,
  url=false,
  doi=false,
  maxbibnames=20,
  maxcitenames=2,
]{biblatex}
\usepackage[dvipsnames]{xcolor}
\usepackage{setspace}
\begin{document}
\title{\textbf{InsurTech innovation using natural language processing}}
\author{Panyi Dong\thanks{Actuarial and Risk Management Sciences, University of Illinois Urbana-Champaign, 1409 W. Green Street (MC-382), Urbana, IL, 61801, USA. Email: \texttt{panyid2@illinois.edu}.}\and Zhiyu Quan\thanks{Actuarial and Risk Management Sciences, University of Illinois Urbana-Champaign, 1409 W. Green Street (MC-382), Urbana, IL, 61801, USA. Email: \texttt{zquan@illinois.edu}.}}
\maketitle

\begin{abstract}

With the rapid rise of InsurTech, traditional insurance companies are increasingly exploring alternative data sources and advanced technologies to sustain their competitive edge. This paper provides both a conceptual overview and practical case studies of natural language processing (NLP) and its emerging applications within insurance operations, focusing on transforming raw, unstructured text into structured data suitable for actuarial analysis and informed decision-making. Leveraging real-world alternative data provided by an InsurTech industry partner that enriches traditional insurance data sources, we apply various NLP techniques to demonstrate feature de-biasing, feature compression, and industry classification in the commercial insurance context. These enriched, text-derived insights not only add to and refine traditional rating factors for commercial insurance pricing but also offer novel perspectives for assessing underlying risk by introducing novel industry classification techniques. Through these industry-proven demonstrations, we show that NLP is not merely a supplementary tool but a foundational element of modern, data-driven insurance analytics. 

\end{abstract}


\section{Introduction}

InsurTech refers to the use of state-of-the-art technology, including both emerging hardware and software, to address inefficiencies across the insurance value chain and further explore new opportunities to reshape traditional business operations. InsurTech encompasses a broad spectrum of technology-driven innovations, including, but not limited to, telematics, usage-based insurance, and the integration of Internet of Things (IoT) sensors. In this study, we focus on a specific class of InsurTech, an InsurTech data vendor, that provides insurance companies with next-generation data solutions. We leverage new and diverse external data sources, such as social media data and online content, to enrich the internal database, thereby empowering actuarial analytics and gaining more accurate insights into risk profiles and policyholder behavior. Specifically, by integrating alternative data sources beyond traditional information, insurance companies can uncover previously unrecognized risk factors, reduce bias in existing features, and identify more accurate risk exposures based on the operational characteristics of the insured entities. This approach allows insurers to move beyond conventional data limitations, enhancing the accuracy and fairness of underwriting and pricing decisions.

Over the past two decades, the insurance industry has increasingly embraced data-driven approaches. Property and Casualty (P\&C) carriers, for instance, began employing predictive modeling in the early 2000s. Since then, insurers have leveraged predictive analytics with large-scale datasets to support a wide range of applications, ranging from risk assessment and ratemaking to fraud detection and litigation forecasting. In the current era of Artificial Intelligence (AI), InsurTech data vendors have become essential partners for insurers striving to maintain competitive advantages. These vendors contribute high-quality external datasets, which are crucial for improving the accuracy and granularity of actuarial models. Social media, in particular, has emerged as one of the most influential categories of alternative data. However, its inherently unstructured and textual nature poses challenges for traditional actuarial modeling frameworks, which are primarily designed to process structured, numerical, or categorical inputs. Furthermore, traditional numerical-only frameworks exhibit several limitations in real-world applications. For example, the numerical rating systems, such as business review scores, provide biased representations of underlying risks due to variations across geographic or socioeconomic groups. In addition, conventional encoding methods for high-dimensional categorical features, such as one-hot encoding, produce sparse and high-dimensional representations while failing to capture semantic relationships among categories.

Processing such massive volumes of unstructured text is infeasible through manual methods alone. The incompatibility between human-readable textual data and numerical-only computer systems presents additional challenges in effectively leveraging the rich sentiment and descriptive information embedded within textual sources. This is where Natural Language Processing (NLP) becomes vital. NLP is an interdisciplinary field that combines machine learning algorithms with computational linguistics to interpret and extract meaning from human language, both written and spoken. In the context of InsurTech vendor data, NLP enables the transformation of raw textual content into structured representations suitable for actuarial modeling and decision-making.

Despite its clear potential and extensive applications in other fields, NLP remains underexplored in actuarial science literature due to limited unstructured textual data availability \citep{Liao2020, Lee2020, zappa_text_2021, xu_bert-based_2022, xu2023framework, sabban2022automatic, Zhang11112024, cao2024assessing}. This study aims to bridge this gap by providing a comprehensive overview of NLP techniques with a focus on their mathematical foundations. We explore essential methods such as sentiment analysis, topic modeling, language models, and embeddings, and discuss their relevance to various insurance applications. In addition, to address the limitations of the numerical-only frameworks in insurance applications discussed above, we propose three insurance application scenarios utilizing the textual data and NLP techniques:
\begin{itemize}
\item Debias existing numerical features by supplementing them with context-aware textual insights;
\item Convert high-cardinality categorical features into dense vector representations;
\item Classify industries and business types based on the textual description.
\end{itemize}
Through empirical studies on real-world InsurTech datasets, we show that NLP is not merely a supplementary tool but a foundational element for modern data-driven insurance analytics. We also highlight promising avenues for future research and development, encouraging the broader actuarial community to explore NLP use cases that extend beyond the scope of this paper. Ultimately, as the insurance industry continues to evolve, we believe NLP will play a central role in driving innovation, improving operational efficiency, and enabling more informed decision-making.

The rest of this paper has been organized as follows. Section \ref{sec:survey} presents a concise overview of key milestones in the development of NLP techniques, serving as an accessible introduction for the actuarial science community. Section \ref{sec:TN} delves into the mathematical foundations of transforming unstructured text into numerical vector representations. In Section \ref{subsec:nlp}, we explore NLP applications in insurance, broadly categorizing them into data enrichment for machine learning tasks and standalone NLP use cases. Finally, Section \ref{subsec:conclude} concludes the paper with key insights and directions for future research.

\section{A Brief Survey of NLP} \label{sec:survey}

Although NLP is considered an emerging field, it is, in fact, not a young discipline. As early as the 1970s, some research related to language processing appeared in research journals. During the 1990s, various statistical methods were already applied to NLP, and some of these methods are still in use today. 

In the book \textit{Foundation of Statistical Natural Language Processing}, \citet{ManningBook} provide a comprehensive overview of statistical NLP methods developed in the 1990s and before, including the n-gram models, Markov Models, and probability-based lexicon building. Additionally, topic modeling techniques such as Latent Semantic Analysis (LSA) \citep{Deerwester1990} and Probabilistic Latent Semantic Analysis (PLSA) \citep{Hofmann2001} were proposed around the same decade, the 1990s. The well-known Recurrent Neural Network (RNN) \citep{elman1990finding} also became popular and remains an important model architecture to this day.

In the 2000s, a widely used topic modeling method, Latent Dirichlet Analysis, was proposed by \citet{Blei2003}. Researchers also focused on representing words as vectors for semantic and syntactic analysis, leading to the development of word embeddings like word2vec \citep{Mikolov2013,Mikolov2013a} and GloVe \citep{Pennington2014GloveGV} between 2012 and 2016. These word embeddings open another level of opportunities to apply natural languages in statistical studies as well as deep learning projects.

In 2017, \citet{Vaswani2017} propose the Attention Mechanism in their paper \textit{Attention Is All You Need} that reshaped the landscape of NLP from RNN to the Transformer architecture. This breakthrough paved the way for the development of GPT \citep{radford2018improving}, BERT \citep{devlin2019}, GPT-2 \citep{radford2019language}, GPT-3 \citep{brown2020language}, GPT-X (later developments), and other pre-trained deep learning networks like Llama \citep{touvronLlamaOpenFoundation2023, grattafioriLlama3Herd2024}, Gemma \citep{teamGemmaOpenModels2024, teamGemma2Improving2024}, Qwen \citep{yangQwen2TechnicalReport2024, wangQwen2VLEnhancingVisionLanguage2024, qwenQwen25TechnicalReport2025} and DeepSeek \citep{deepseek-aiDeepSeekV2StrongEconomical2024, deepseek-aiDeepSeekV3TechnicalReport2024, guoDeepSeekR1IncentivizesReasoning2025}, which have pushed NLP to unprecedented levels of performance and ushered in the era of ChatGPT. 

In addition, there are alternative architectures that differ from the Transformer and may have the potential to further advance Large Language Models (LLMs), to list a few, State Space Models (SSMs) like Mamba \citep{guMambaLinearTimeSequence2024}, and newly designed Recurrent Models like RWKV \citep{pengRWKVReinventingRNNs2023}.

Due to the rapid pace of advancements in this field, we have highlighted only a selection of representative open-source studies to serve as a foundation for further academic exploration. The studies referenced above provide a sketch of the evolution of mainstream NLP, offering a foundation relevant to this paper. Given the rapid advancements in NLP research, this sketch may already be outdated by the time of publication. However, this paper still serves as the first attempt to introduce systematic and comprehensive NLP literature to the actuarial science field. In the subsequent sections, we explore various NLP techniques in depth and examine their applicability to different areas within actuarial science.

\section{Text to Numbers} \label{sec:TN}

Textual data requires special preparation prior to its input into a computing system, and the process of transforming it into numerical representations constitutes a crucial aspect of NLP.

\subsection{Text Preprocessing}

In any domain, data requires preprocessing, and this holds true for textual data as well. Typically, textual data preprocessing follows a standard procedure comprising two primary steps: Text Cleaning and Lexicon Normalization.

\subsubsection{Text Cleaning} \label{subsubsec:clean}

Text cleaning is akin to eliminating noise. In essence, it involves the removal of text segments that lack relevance to the context of the data or the eventual output. Though time-consuming, this process is crucial for effectively handling real-world text data. Despite recent advancements in transformer architectures and other methods driven by massive data and heavy computation, text cleaning still remains vital.

The basic text cleaning is applied at the initial stage of the entire model pipeline, encompassing, but not limited to: Contractions, Letter case normalization, Filtering, and Stopwords removal. For additional details on each technique and its implications for the insurance domain, please refer to Appendix \ref{appendix_sec:cleaning}. After these initial text cleaning steps, the refined content is ready for the subsequent phase of text preprocessing.

\subsubsection{Lexicon Normalization: Tokenizing, Stemming, Lemmatization} \label{subsubsec:tsl}

In the subsequent phase of text preprocessing, various linguistic techniques are applied to the corpus to normalize the texts. \textit{Tokenizing} involves separating the corpus into words or phrases that are treated as tokens. In some cases, these tokens can be encoded into numerical representations, allowing for the numeric representation of sentences, paragraphs, and even entire documents through further preprocessing. The collection of distinct tokens is referred to as the \textit{dictionary} of a document collection. To minimize the size of the dictionary, techniques such as Stemming and Lemmatization are applied. \textit{Stemming} involves obtaining the root form of words. Usually, a word has multiple forms, such as the plural `s' of nouns, the `ing' of verbs, or other affixes, and reducing these forms to a common root simplifies the complexity of the corpus while preserving meaning. The widely used algorithm, the Porter-Stemmer, was introduced in 1980 by \citet{Porter1980}, which employs a set of production rules for the iterative transformation of English words into their stems. \textit{Lemmatization} is another method that can be applied to the corpus to reduce complexity by mapping verb forms to the infinitive tense and nouns to the singular form. Lemmatization uses the morphological information of each word to simplify the word into its base form.

\subsection{Embedding}  \label{subsec:embedding}

The text preprocessing in Section \ref{subsubsec:clean} and Section \ref{subsubsec:tsl} pertains to fundamental preprocessing. For years, researchers have sought to represent words numerically to make text data more amenable to numerical analysis. While tokenizing enables the encoding of sentences with numbers through one-hot encoding for each word, this approach results in extremely high-dimensional vectors and a loss of semantic meaning in words, given the nature of textual data. \textit{Word Embedding} addresses this challenge by embedding words into numerical representations in a concise and meaningful manner for further analysis. This is particularly crucial in NLP, where word vector representations not only significantly reduce corpus dimensionality but also facilitate the mathematical and computational analysis of natural languages.

The exploration of word embedding commenced with the Vector Space Model \citep{Salton1975} in 1975. In 1988, a method that uses Latent Semantic Analysis was proposed by \citet{Deerwester1990}. Subsequently, in the early 2000s, \citet{Bengio2001} invented the \textit{embedding layer} in their Neural Network Language Model paper. Then, word embedding relies on contextual information and word relationships, with influential approaches like word2vec \citep{Mikolov2013, Mikolov2013a}, GloVe \citep{Pennington2014GloveGV}, GPT \citep{radford2018improving}, BERT \citep{devlin2019} and its variants. These methods have revolutionized word embedding and found applications beyond NLP, extending to fields like computer vision and recommender systems development.

\subsection{Vector Space Model}
\label{subsubsec:TF-IDF}

The Vector Space Model holds significance in NLP despite its age, since it is the first proposal to embed words into vectors that allows the algebraic representation for textual data. Further, the vector representation of terms serves as the cornerstone for incorporating textual data into deep learning models. \citet{Salton1975} propose representing a document \textit{$d_j$}, $j=1, \dots, n$, using an \textit{m-dimensional} vector with selected terms \textit{$t_i$}, $i=1, \dots, m$, within a corpus \textit{$D$} containing $n$ documents. The definition of a term, often dictated by the application, can encompass single words, keywords, or longer phrases. The value within the vector is the weight of each term in the document with terms weighted based on their importance. \citet{Salton1975} employ the Term Frequency Inverse Document Frequency (TF-IDF) method to weigh the terms within the document. The mathematical formulation of TF-IDF is articulated through a series of equations that define its components and provide an intuitive understanding. The Term Frequency (TF) refers to the relative frequency of a specific term within a document and can be represented as 
\begin{equation}
    TF(t,d) = \frac{N_{td}}{N_d}
\end{equation}
where $N_{td}$ is the number of times term $t$ appears in document $d$ and $N_d$ is the total number of terms in document $d$. Here, we omit subscripts for simplicity. The Inverse Document Frequency (IDF) quantifies the amount of information conveyed by a term, indicating whether it is common or rare across all documents. It is given by 
\begin{equation}
    IDF(t, D) = log(\frac{n}{N_t})
\end{equation}
where $n$ is the number of documents in corpus $D$ and $N_t$ is the number of documents contains term $t$. The TF-IDF value is then calculated as:
\begin{equation}
    TF-IDF(t, d, D) = TF(t,d) \times IDF(t, D)
\end{equation}

There are multiple ways to define $TF(t,d)$ and $IDF(t, D)$, along with their combined weighting schemes $TF-IDF(t, d, D)$, each grounded in different intuitions for representing term weight within a document. Additionally, Positive Pointwise Mutual Information (PPMI) offers an alternative scheme of term weighting using term-term matrices, where vector dimensions correspond to terms rather than documents. PPMI draws on the intuition that the strength of association between two terms should be measured by how much more frequently they co-occur in a corpus than would be expected by chance. By leveraging pointwise mutual information, which quantifies this co-occurrence relative to an independence assumption, \citet{church1990word} define alternative intuitions to represent term weight within surrounding terms. These intuitions provide valuable insights that enhance the interpretability of NLP models. This interpretability is particularly critical in the field of insurance, where understanding the rationale behind model predictions is essential for risk assessment, regulatory compliance, and building trust with stakeholders. By ensuring that the term weighting schemes and their combinations are intuitively grounded, insurance practitioners can better understand and explain the model's decision-making processes.

\subsection{Statistical Language Model}
\label{subsubsec:LSA}

As discussed in the previous section, documents can be represented as vectors of terms in the Vector Space Model. However, a significant drawback of this approach is that it can suffer from high-dimensionality, particularly when dealing with a large corpus. To address this issue, \citet{Deerwester1990} propose Latent Semantic Analysis (LSA), which assumes that terms with similar meanings usually appear in similar contexts within documents. Based on this assumption, LSA analyzes relationships between a set of documents and their terms by producing a set of concepts that capture the underlying structure of the documents and terms.

To be more specific, LSA constructs a term-document occurrence matrix $A_D$, where rows represent unique terms and columns represent each document, capturing the term counts per document. 
$$
A_D = 
    \begin{bmatrix}
    a_{11} & \cdots & a_{1n} \\
    \vdots & \ddots & \vdots \\
    a_{m1} & \cdots & a_{mn}
    \end{bmatrix}
$$ where $a_{ij}$ being the frequency of the selected term $t_i$, $i=1,\dots,m$, in document $d_j$, $j=1,\dots,n$, within the corpus \textit{D}. The choice of $a_{ij}$ can be a straightforward frequency count or employ more sophisticated methods such as $TF-IDF(t_i, d_j, D)$. When $n$ and $m$ are large and $A_D$ is sparse, it is beneficial to use matrix factorization techniques, such as Singular Value Decomposition (SVD), to reduce the dimensionality of the term-document occurrence matrix while preserving the similarity structure, thereby uncovering latent structures within the data.

Using SVD, the term-document occurrence matrix $A_D$ can be decomposed as
\begin{equation}
A_D = U\Sigma V^T
\end{equation}
where the columns of $U$ and $V$ represent the term vectors and document vectors respectively, and $\Sigma$ is a diagonal matrix with singular values. By retaining only the $k$ (where $k < min(m, n)$) largest singular values and their associated vectors from $U$ and $V$, we can approximate $A_D$ by $A_k$ with minimal error.
\begin{equation}
A_D \approx A_k = U_k \Sigma_k V_k^T
\end{equation}
Therefore, $m$ rows, vectors corresponding to a term, in $A_D$ map to $k$ columns of $U_k$, which represents a lower-dimensional space. Similarly, the $n$ columns, vectors corresponding to a document, in $A_D$ map to $k$ columns of $V_k$. This lower-dimensional approximation condenses the higher-dimensional space by retaining essential information and filtering out incidental occurrences of terms and documents. In the original matrix $A_D$, each entry represents each term presented in each document. In contrast, in the lower-dimensional representation, only terms related to each document are considered in $U_k$. However, it is worth pointing out that these new dimensions do not correspond to comprehensible concepts and, therefore, lose interpretability. In addition, the lower-dimensional approximation is derived purely through mathematical matrix operations without incorporating any consideration of natural language structure or linguistic meaning. Inspired by the early SVD-based approaches, several alternative matrix factorization approaches have been developed, including Probabilistic Latent Semantic Indexing (PLSI) \citep{10.1145/312624.312649}, Latent Dirichlet Allocation (LDA) \citep{Blei2003}, and Non-negative Matrix Factorization (NMF) \citep{lee1999learning}, each introducing different probabilistic and structural assumptions to improve representation quality.

SVD and LSA attempt to establish relationships between terms and documents; however, they overlook the relationships between terms themselves, particularly the order in which terms are presented in the document. Another approach, known as a Language Model (LM), involves representing terms probabilistically to predict a term based on the preceding terms. For example, the $\mathcal{N}$-\textit{gram language model} predicts a term given the $\mathcal{N}-1$ term, thereby assigning probabilities to entire $\mathcal{N}$ sequences. The $\mathcal{N}$-gram LM is developed based on the idea from \citet{markov1913essai}, known as Markov chains, to predict the next word.

The Markov-based LM can be mathematically formulated using the conditional probabilities of terms given their preceding context within a document. Assume a term $t_i$ is a single word and $i$, $i=1,\dots,m$, preserves the natural language order in a document $d_j$, $j=1,\dots,n$, we can formulate the joint probability of each word in a document having a particular value
$$
\begin{aligned}
P\left(d_j\right) & = P\left(X_1=t_1, X_2=t_2, \ldots, X_m=t_m\right) \\
& = P\left(X_1=t_1\right) P\left(X_2=t_2 \mid X_1=t_1\right) P\left(X_3=t_3 \mid X_1=t_1, X_2=t_2\right) \ldots \\
&P\left(X_m = t_m \mid X_1=t_1, X_2=t_2, \ldots, X_{m-1}=t_{m-1}\right) \\
& =\prod_{i=1}^m P\left(X_i = t_i \mid X_1=t_1, X_2=t_2, \ldots, X_{i-1}=t_{i-1}\right)
\end{aligned}
$$
It is computationally infeasible to determine $P\left(X_i = t_i \mid X_1=t_1, X_2=t_2, \ldots, X_{i-1}=t_{i-1}\right)$ or the underlying corpus may not be sufficient to provide meaningful probabilities. The \textit{bigram language model} approximates the probability of a word given all the previous words $P\left(X_i = t_i \mid X_1=t_1, X_2=t_2, \ldots, X_{i-1}=t_{i-1}\right)$ by using only the conditional probability of the preceding word $P\left(X_i = t_i \mid X_{i-1}=t_{i-1}\right)$. Obviously, the probability of the document is $P\left(d_j\right) \approx \prod_{i=1}^m P\left(X_i = t_i \mid X_{i-1}=t_{i-1}\right)$. The approximation arises from the assumption that the probability of a word depends only on the previous word, known as the Markov assumption. Markov models, a class of probabilistic models, operate under the premise that the probability of a future word can be predicted without considering too many past words. We can generalize the bigram to the $\mathcal{N}$-gram LM following
\begin{equation}
P\left(X_i = t_i \mid X_1=t_1, X_2=t_2, \ldots, X_{i-1}=t_{i-1}\right) \approx P\left(X_i = t_i \mid X_{i-\mathcal{N}+1}=t_{i-\mathcal{N}+1}, \ldots, X_{i-1}=t_{i-1}\right)
\label{eq:objectprob}
\end{equation}

An intuitive way to estimate these probabilities is through Maximum Likelihood Estimation (MLE). We obtain the MLE estimates for the parameters of an $\mathcal{N}$-gram model by counting occurrences in a corpus, and normalizing these counts to lie between $0$ and $1$.

The definition of probability and the number of past words to consider in an $\mathcal{N}$-gram model can be guided by domain-specific knowledge, which plays a critical role in achieving meaningful and accurate results. This is particularly true in specialized fields such as the insurance domain, where the language used often contains technical jargon, standardized phrases, and specific terminologies. For instance, in insurance documents, certain terms may frequently co-occur or follow specific sequences due to contractual obligations or regulatory requirements. By incorporating domain knowledge, the $\mathcal{N}$-gram model can be tailored to prioritize these patterns, ensuring that the probability estimations align more closely with real-world use cases and improve language models within the insurance context.

Despite the remarkably simple and clear formalization of $\mathcal{N}$-gram LM, it faces significant challenges, primarily due to the exponential increase in the number of parameters as the $\mathcal{N}$-gram order increases. Modern LM often employs Neural Networks (NN), which address this issue by projecting words into a continuous space where words with similar contexts have similar representations. 

\subsection{Neural Language Model}

As mentioned in the previous section, the idea of vector semantics is to represent a term as a point in a multidimensional semantic space. An alternative way to find vector representation is through the distributions of word neighbors. The word2vec model family utilizes either of two model architectures: Continuous Bag-Of-Words (CBOW) or continuously sliding skip-gram, to construct short, dense vectors with useful semantic properties. The skip-gram approach ensures that the vector of a term is close to the vector of each of its neighbors, while CBOW ensures that the vector-sum of a term's neighbors is close to the vector of the term. In both architectures, word2vec considers both individual terms and a sliding context window as it iterates over the corpus.
 
Assume a term $t_i$ is a single word and index $i$, $i=1,\dots,m$, preserves the natural language order in a document $d_j$, $j=1,\dots,n$, within a corpus \textit{$D$}. For context window $C$, we consider the $2C$ neighboring terms, $t_{i-C}$, $t_{i-C+1}$, $\dots$, $t_{i-1}$, $t_{i+1}$, $\dots$, $t_{i+C}$ which form the neighboring context for the term $t_i$. In CBOW, based on the neighboring context, we can maximize the following probability
\begin{equation}
\prod_{i \in D}  P\left(X_i = t_i \mid X_{i-C}=t_{i-C}, \dots, X_{i-1}=t_{i-1}, X_{i+1}=t_{i+1}, \dots, X_{i+C}=t_{i+C}\right)
\label{eq:objectcbow}
\end{equation}
which is equivalent to maximizing the log-probability
\begin{equation}
\sum_{i \in D} ln P\left(X_i = t_i \mid X_{i-C}=t_{i-C}, \dots, X_{i-1}=t_{i-1}, X_{i+1}=t_{i+1}, \dots, X_{i+C}=t_{i+C}\right)
\label{eq:logprob}
\end{equation}
We can approach this optimization problem using MLE as mentioned in the context of n-gram LM. However, there is an alternative approach with word2vec, which serves as a bridge between statistical methods and NN (specifically, shallow NN with a sigmoid activation function).

For the vector representation, let $v_{t_i}$ denote the vector corresponding to the term $t_i$, and $v_{C}$ is the sum of the neighboring context, i.e., $v_{C} = v_{t_{i-C}}+\dots+v_{t_{i-1}}+v_{t_{i+1}}+\dots+v_{t_{i+C}}$. If we use the vector product as a measure of term similarity, we can further normalize the vector product using the logistic or sigmoid function to induce the probability.
$$
P\left(X_i = t_i \mid X_{i-C}=t_{i-C}, \dots, X_{i-1}=t_{i-1}, X_{i+1}=t_{i+1},, \dots, X_{i+C}=t_{i+C}\right)= \frac{e^{v_{t_i} \cdot v_{C}}}{\sum_{i \in D} e^{v_{t_i} \cdot v_{C}}}
$$
The probability is designed to be higher when the term is in its neighboring context. Then we can specify Expression \ref{eq:logprob} as follows:
\begin{equation}
\sum_{i \in D}\left(v_{t_i} \cdot v_{C}-\ln \sum_{i \in D} e^{v_{t_i} \cdot v_{C}}\right)
\label{eq:vecprob}
\end{equation}
To find the $v_{t_i}$ that maximizes the objective function Expression \ref{eq:vecprob}, we can take two different approaches. The first is using a logistic regression setting, which creates a binary response variable based on whether the terms belong to the neighborhood or not. This involves negative sampling, where terms that are outside the neighboring context within the corpus are randomly chosen. The second approach utilizes shallow NN to find the best representation.

In the continuously sliding skip-gram model, instead of using the neighboring context to predict the term, the approach is reversed, using the term to predict its neighboring context terms. Hence, the objective function, which reverses the conditional probability in Expression \ref{eq:objectcbow}, is formulated for maximization as follows.
\begin{equation}
\prod_{i \in D}  P\left( X_{i-C}=t_{i-C}, \dots, X_{i-1}=t_{i-1}, X_{i+1}=t_{i+1}, \dots, X_{i+C}=t_{i+C} \mid  X_i = t_i \right)
\label{eq:prob}
\end{equation}
With the assumption of independence, i.e., nonadjacent words are independent, we can simplify Expression \ref{eq:prob} as follows:
\begin{equation}
\prod_{i \in D} \prod_{\substack{c=i-C \\ c\neq i}}^{i+C}  P\left( X_{c}=t_{c} \mid  X_i = t_i \right)
\end{equation}
The remaining steps are very similar to those discussed in the CBOW method.

Word2vec embeddings are considered \textit{static embeddings}, which means that they learn a fixed representation for each term in the corpus. Several other static embedding models have been developed to address specific challenges. For example, fastText\footnote{FastText is an open-source library that has 157 languages pre-trained embeddings. See \href{https://fasttext.cc/}{https://fasttext.cc/}} \citep{bojanowski-etal-2017-enriching}, an extension of word2vec, tackles the issue of unknown terms—those that appear in a corpus but are unseen during training. Additionally, it addresses word sparsity, where infrequent variations of nouns and verbs may have limited occurrences. FastText overcomes these problems by incorporating subword models, representing each term as a combination of itself and a bag of constituent $\mathcal{N}$-grams, with special boundary symbols added. This allows unknown terms to be represented solely by the sum of their constituent $\mathcal{N}$-grams.

Another widely used static embedding model is GloVe \citep{Pennington2014GloveGV}, which captures global corpus statistics. Unlike word2vec, which relies on local context windows, GloVe constructs word embeddings based on ratios of probabilities derived from the term-term co-occurrence matrix. This approach effectively combines the strengths of count-based models like PPMI, which leverage corpus-wide statistics, and predictive models like word2vec, which capture linear relationships between terms.

\begin{figure}[h!]
\centering
\subfigure[Feedforward NN Neural LM]{\includegraphics[scale=0.18]{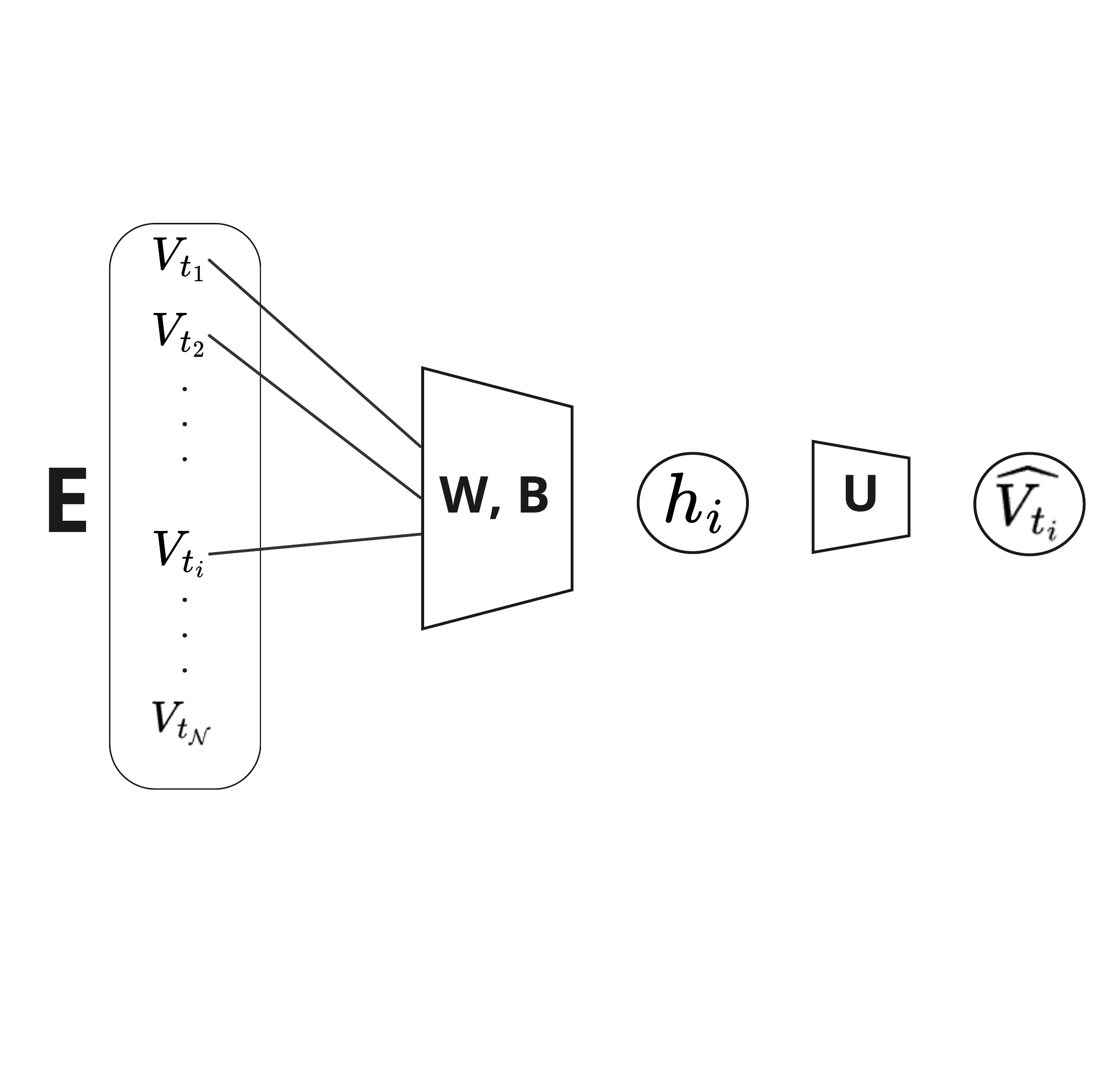}\label{fig:fnn}}
\subfigure[RNN Neural LM]{\includegraphics[scale=0.22]{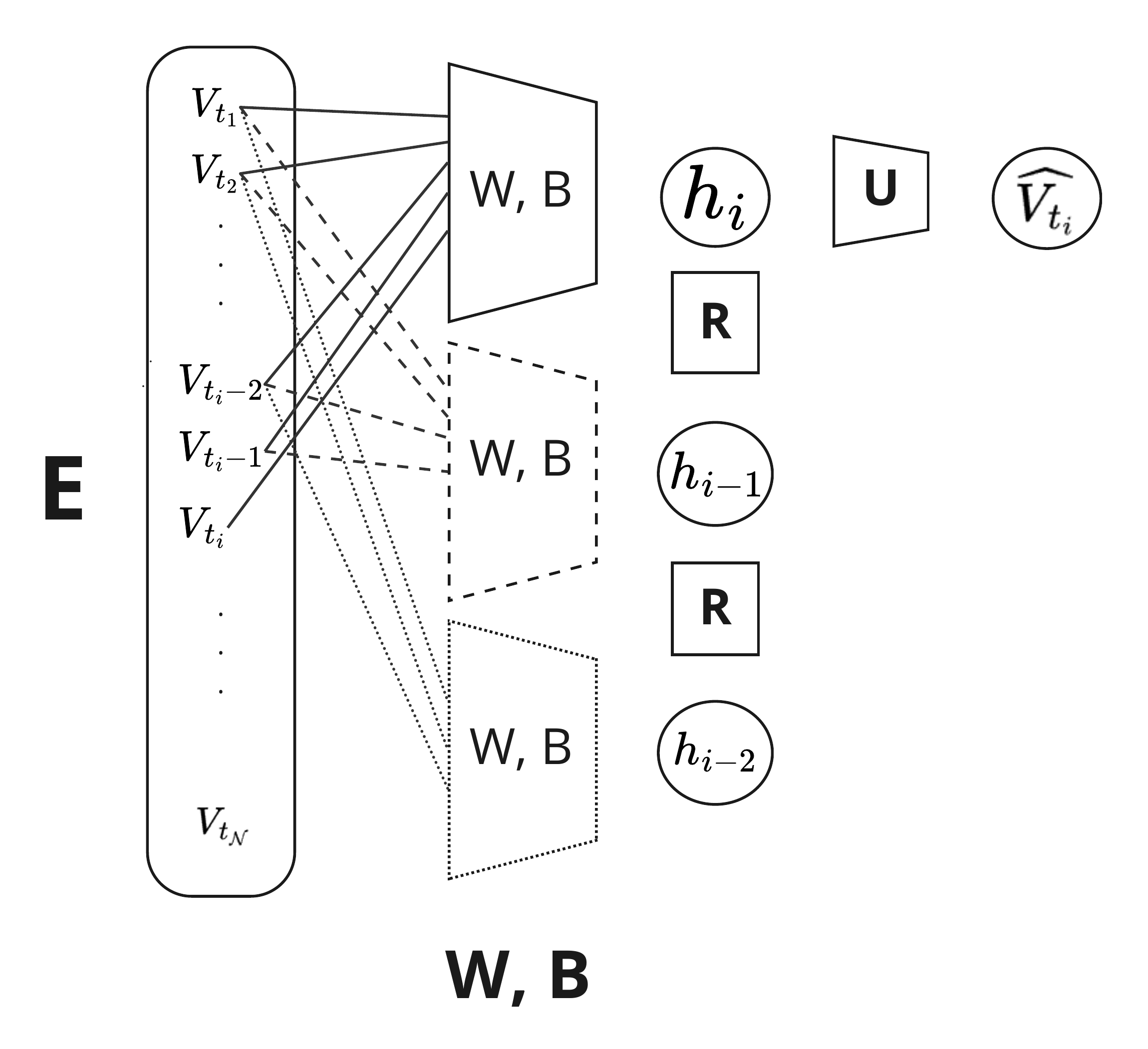}\label{fig:rnn}}
\caption{Neural LM Architectures}
\label{fig:lm}
\end{figure}

\citet{Bengio2001} is one of the pioneers in introducing neural LM using a feedforward NN. The objective function remains the same as in the $\mathcal{N}$-gram LM, as described in Expression \ref{eq:objectprob}. We denote the feedforward NN as $f(\textbf{E}, \textbf{W}, \textbf{b}, \textbf{U})$, where $\textbf{E}$ represents the embeddings for the input terms; precisely, the matrix of the concatenated $\mathcal{N}$ previous terms' vector representation $v_{t_i}, i=1,\dots, \mathcal{N}$. These embeddings can be sourced from word2vec, GloVe, or any other pre-trained embedding models. The parameters $\textbf{W}$ and $\textbf{b}$ correspond to the conventional weight matrix and bias vector of the feedforward NN. Additionally, to distinguish the different neural LM architectures, we introduce the weight matrix $\textbf{U}$, which connects the hidden layers, $h$, to the output layer. In traditional feedforward NN, $\textbf{U}$ is typically a part of $\textbf{W}$. Given an appropriate loss function, the feedforward NN is trained to predict the next term based on the $\mathcal{N}$ preceding input terms. This training process not only results in an LM, but also produces a new set of optimized embeddings $\textbf{E}$, which can serve as improved word representations for the training corpus. Figure \ref{fig:fnn} depicts the architecture of the neural LM using a simple feedforward NN.

Language is inherently sequential, unfolding over time in both written and spoken forms. However, feedforward NN lacks a built-in mechanism to capture this temporal dependency. An alternative deep learning architecture that explicitly models sequential dependencies is the Recurrent Neural Network (RNN) \citep{elman1990finding}, along with its more advanced variants, such as Long Short-Term Memory (LSTM) networks \citep{10.1162/neco.1997.9.8.1735}.

RNN is characterized by cyclic connections within its network, allowing it to retain information from previous time steps, i.e., the value of some unit is directly or indirectly dependent on its own earlier outputs as an input. In contrast to feedforward NN, where input flows strictly in one direction, RNN introduces an additional weight matrix $\textbf{R}$ that connects hidden layers generated across sequential inputs (see Figure \ref{fig:rnn}). This recurrent architecture enables the model to capture long-range dependencies, allowing decisions to be influenced by context, extending hundreds of terms back. Unlike $\mathcal{N}$-gram LM, which is constrained by a fixed-length context, and feedforward LM, which operates with a static context window, RNN maintains a dynamic hidden state that, in principle, can encode information from the entire preceding sequence.

The flexibility of RNN has led to the development of more sophisticated architectures. Stacked RNN consists of multiple layers of simple RNNs, where the output of one layer serves as input to the next, enabling the model to capture more complex hierarchical dependencies. Bidirectional RNN (BiRNN) \citep{650093} further enhances performance by incorporating two independent RNNs, one processing the sequence from start to end and the other in reverse, thus leveraging both past and future context, similar to how humans read and comprehend text by revisiting previous sections or jumping to the end as needed. BiRNN is particularly useful for situations where the entire input sequence is available and allows access to information from both directions, such as sentiment analysis or question answering. LSTM addresses a key limitation of traditional RNN: the challenge of retaining relevant information over long sequences. Standard RNN struggles with \textit{vanishing gradients}, where long-term semantic information is gradually lost due to repeated gradient multiplication, making it difficult to preserve context across many time steps. LSTM mitigates this issue by introducing an explicit memory cell that allows the model to learn what information to retain and what to discard. The core mechanism behind LSTM is adding an explicit context layer to the architecture and the use of gating functions—add, forget, and output gates—that regulate the flow of information into, within, and out of the memory cell. These gates are governed by additional weight parameters, which enable the network to dynamically update its internal state based on sequential inputs. By learning to selectively remember and forget information, LSTM significantly enhances the model's ability to capture long-term dependencies and improve language modeling performance.

Expanding on the RNN model architecture, encoder-decoder networks—also known as sequence-to-sequence models—are designed to generate contextually appropriate output sequences of arbitrary length given an input sequence. The fundamental idea behind encoder-decoder networks is the use of an encoder network to process the input sequence and create a contextualized representation, often referred to as the \textit{context vector}. This representation is then passed to a decoder, which generates a task-specific output sequence. The modern neural encoder-decoder architecture is pioneered by \citet{kalchbrenner-blunsom-2013-recurrent}, who introduced a convolutional neural network (CNN) encoder paired with an RNN decoder. \citet{cho-etal-2014-learning}, who coined the term encoder-decoder, and \citet{NIPS2014_a14ac55a} show how to use extended RNNs for both encoder and decoder. A significant advancement in encoder-decoder networks comes with the introduction of the \textit{attention mechanism}. The concept of providing the generative decoder with a weighted combination of input representations rather than relying solely on a fixed-size context vector was first explored by \citet{graves2013generating} in the context of handwriting recognition. The primary function of attention is to alleviate the bottleneck problem by allowing the decoder to access all hidden states of the encoder rather than just the final one. At its core, attention is a weighted sum of context vectors, with additional mechanisms for computing the weights dynamically based on input relevance.

Building upon attention mechanisms, the \textit{transformer} architecture \citep{Vaswani2017} introduced a paradigm shift by eliminating recurrence entirely. Unlike RNNs, which process sequences step-by-step, transformers leverage attention mechanisms to model dependencies between terms in parallel, significantly improving efficiency and scalability. A key advantage of transformers is their ability to generate contextual embeddings, where the representation of a term is dynamically adjusted based on its surrounding context, unlike static embeddings such as word2vec, which assign a fixed vector to each word regardless of its usage. The input representation in transformers is more complex compared to RNNs, as it involves separately computing two embeddings: (1) an input token embedding, which represents the meaning of the term, and (2) an input positional embedding, which encodes the term's position within the sequence. This additional positional information is crucial since transformers, unlike RNNs, do not inherently process input in sequential order. The combination of these embeddings, along with multi-head attention and feedforward layers, allows transformers to capture nuanced contextual relationships, making them the backbone of modern NLP models such as BERT \citep{devlin2019} and GPT \citep{brown2020language}. At the time of writing this paper, the most recent advancements in NLP largely remain based on encoder-decoder architectures and their variants.

\begin{figure}
\centering
\includegraphics[width=\linewidth]{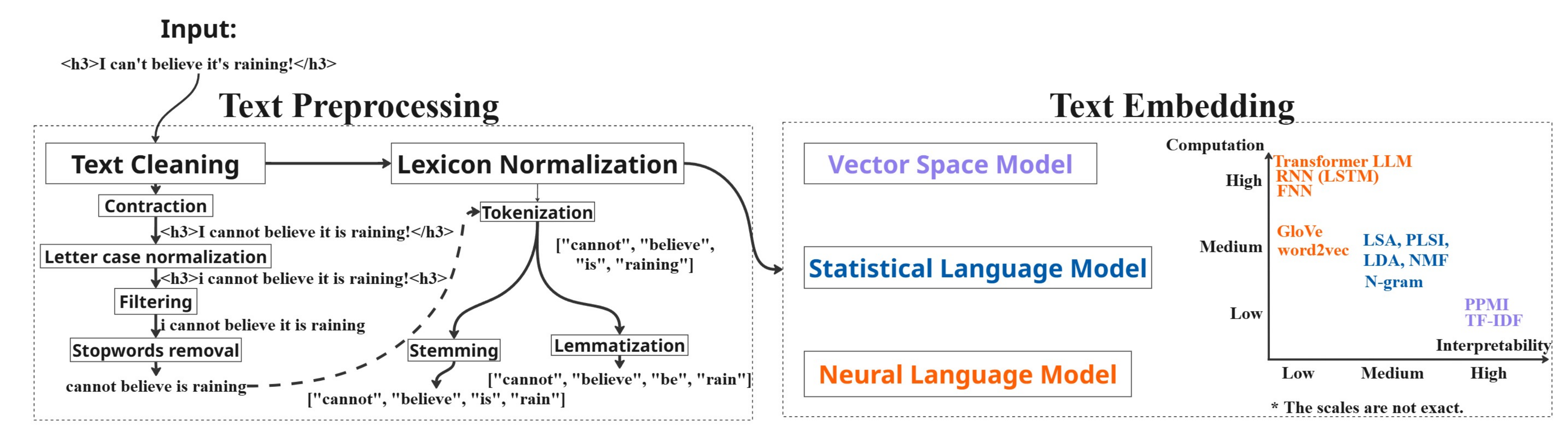}
\caption{Summary of Text to Numbers Solutions}
\label{fig:text-to-number}
\end{figure}

Neural LM offers several advantages over $\mathcal{N}$-gram LM. Neural LM can capture much longer contextual dependencies, generalize more effectively across similar word contexts, and achieve higher accuracy in word prediction. However, these benefits come at a cost. As depicted in Figure \ref{fig:text-to-number}, neural LMs are significantly more complex, require greater computational resources and energy for training, and tend to be less interpretable than $\mathcal{N}$-gram models. As a result, for certain smaller-scale tasks or insurance applications where interpretability and efficiency are prioritized, traditional $\mathcal{N}$-gram models may still be the preferred choice. This is why we introduce comprehensive NLP literature that potentially pertains to actuarial science.  In Figure \ref{fig:text-to-number}, we summarize the text-to-numbers workflow and list the referenced NLP models mentioned in the previous discussion.

\section{NLP in Action: Use-cases Leveraging InsurTech Datasets} \label{subsec:nlp}

We implement NLP techniques within the business insurance domain using real-world datasets obtained from Carpe Data\footnote{\href{https://carpe.io}{https://carpe.io}}. In our analysis, we categorize NLP applications into two key areas: data enrichment and standalone use cases. 
Data enrichment refers to coupling extracted textual information with classical or innovative modeling techniques to strengthen traditional use cases, while standalone use cases investigate entirely new opportunities that arise specifically from the integration of textual data. Table \ref{tab:nlp-in-insurance} summarizes existing insurance NLP-related solutions in these two key areas. Furthermore, the table underscores our contributions to NLP solutions in the insurance sector, where we leverage proprietary industry data with up-to-date NLP techniques, explore novel real-world use cases, and provide comprehensive coverage across data enrichment and standalone applications. By leveraging InsurTech datasets, we showcase how NLP transforms raw textual data into actionable intelligence, unlocking new opportunities for insurers.

\begin{table}[!ht]
\scriptsize
\centering
\begin{tabular}{cccc}
\toprule
Reference & Text Embedding Technique & Use Case & Key Area \\
\midrule
\citet{Lee2020} & word2vec/GloVe & Loss Modeling & Data Enrichment \\[0.2ex]
\citet{manski_extracting_2021} & GloVe & Loss Modeling & Data Enrichment \\[0.2ex]
\citet{baillargeon2021mining} & TF-IDF $\mathcal{N}$-grams/GloVe & Loss Modeling & Data Enrichment \\[0.2ex]
\citet{zappa_text_2021} & $\mathcal{N}$-gram & Information Retrieval & Data Enrichment \\[0.2ex]
\citet{manski_loss_2022} & word2vec & Loss Modeling & Data Enrichment \\[0.2ex]
\citet{xu_bert-based_2022} & BERT & Loss Modeling & Data Enrichment \\[0.2ex]
\citet{sabban2022automatic} & fastText & Loss Modeling & Data Enrichment \\[0.2ex]
\citet{xu2023framework} & BERT & Loss Modeling & Data Enrichment \\[0.2ex]
\citet{troxler2024actuarial} & DistilBERT & Loss Modeling & Data Enrichment \\[0.2ex]

Feature De-Biasing (ours) & WordNet & Sentiment Analysis & Data Enrichment \\[0.2ex]
Feature Compression (ours) & BGE & Loss Modeling & Data Enrichment \\[0.2ex]

\citet{Liao2020} & TF-IDF & Customer Service & Standalone Application \\[0.2ex]
\multirow{3}*{\citet{Zhang11112024}} & \multirow{3}*{Sentence-BERT} & Information Retrieval & \multirow{3}*{Standalone Application} \\[0.2ex]
& & Text Classification & \\[0.2ex]
& & Documentation Summarization & \\[0.2ex]
\citet{cao2024assessing} & BERT/BGE/LLaMA3 & Text Classification & Standalone Application \\[0.2ex]
\citet{liTextualAnalysisInsurance2025} & LLM & Sentiment Analysis & Standalone Application \\[0.2ex]

Industry Classification (ours) & LLM & Text Classification & Standalone Application \\[0.2ex]
\bottomrule
\end{tabular}
\caption{Overview of Insurance NLP-related Solutions}
\label{tab:nlp-in-insurance}
\end{table}

\subsection{InsurTech Data}

InsurTech, much like FinTech, has emerged as a key catalyst in transforming the conventional insurance sector into a modern data-driven industry in the era of big data and AI. This transformation introduces a diversity of data sources and formats, among which textual data has become one of the most prominent and extensively studied in the current AI landscape.
A vast amount of textual data can be scraped from the internet, including information that businesses provide about their products and services on their websites, as well as customer reviews and feedback shared on social media and consumer rating platforms. While this data is relatively easy to obtain, extracting meaningful insights from it remains a challenge. However, advancements in NLP techniques offer powerful tools to analyze this unstructured text effectively. These methods allow businesses to gain a deeper understanding of their operations and customers' perceptions of their products and services. For insurers, leveraging these insights can enhance risk management by identifying potential vulnerabilities, improving customer segmentation, and refining underwriting and claims processes. 
In this work, we extend the real-world InsurTech dataset introduced in \citet{Quan15102024}, by incorporating additional textual data to support our feature de-biasing and industry classification use cases. 
We obtained the raw data from our InsurTech partner and applied a series of data preprocessing steps, as discussed in Section \ref{sec:TN}, to prepare the data for downstream predictive modeling.
Since our study incorporates only a limited set of numerical variables in \citet{Quan15102024}, and the majority of our NLP use cases rely exclusively on newly introduced textual data, we omit unnecessary details of irrelevant numerical variables and instead discuss the adopted numerical features within the context of the corresponding use cases.
To demonstrate the value of textual data in enhancing insurance operations, we specifically incorporate business reviews scraped from crowd-sourced platforms (e.g., Google Maps and Yelp user reviews) as well as descriptive texts provided by insured businesses on their websites. 
Regarding the textual components, since the InsurTech data is proprietary and the majority of the textual information contains easily identifiable details about the underlying businesses, we are unable to present the raw text directly. 
Nevertheless, in our use cases, we provide selected examples that illustrate the data while ensuring that no sensitive information is disclosed. 
Furthermore, to facilitate reproducibility and encourage adoption in related applications, we release the review dataset and the synthetic industry classification dataset used in Subsections \ref{subsubsec:senti} and \ref{subsubsec:class}, respectively, as well as the source code for all experiments.\footnote{\href{https://github.com/PanyiDong/InsurTech_NLP}{https://github.com/PanyiDong/InsurTech\_NLP}}

\subsection{Data Enrichment for Machine Learning Tasks} \label{subsec:data}

NLP can significantly enhance existing insurance datasets by extracting valuable insights from unstructured text sources, such as customer reviews, claim descriptions, policy documents, and even social media posts. By converting this unstructured data into structured information, NLP enables insurers to uncover hidden patterns and identify additional risk factors that would otherwise remain undetected. This data enrichment process not only broadens the scope of available features for predictive modeling but also improves the accuracy of risk assessment and the efficiency of operations, e.g., improving customer experience and underwriting decisions. 

\subsubsection{Data Enrichment Using Text} \label{subsubsec:preact}

As discussed extensively in Section \ref{sec:TN}, converting textual data into numerical representations is a well-established tradition that continues to evolve. In actuarial science, numerous efforts have been made to leverage textual data to enhance predictive modeling in insurance. For instance, \citet{Lee2020} introduce a framework that incorporates textual data into insurance claims modeling and examine its applications in claims management processes. Building on this foundation, \citet{manski_extracting_2021} and \citet{manski_loss_2022} present a series of studies that utilize embedding techniques to enhance pricing models. Similarly, \citet{zappa_text_2021} demonstrate the application of NLP techniques in insurance by analyzing accident narratives from police reports to classify risk profiles and fine-tune policy premiums. 
\citet{baillargeon2021mining} propose a data-driven approach that automatically extracts actuarial risk factors from a large corpus of accident descriptions, further emphasizing the potential of NLP in refining insurance models. 
Expanding on these efforts, \citet{xu_bert-based_2022} and \citet{xu2023framework} employ BERT embeddings to enhance the frequency and severity prediction of truck warranty claims, underscoring the transformative impact of NLP in improving predictive models for actuarial tasks. \citet{sabban2022automatic} extend the use of text-based claim predictions to imbalanced datasets by developing a framework that leverages neural networks to automatically identify extremely severe claims. 
\citet{troxler2024actuarial} embed car accident descriptions into condensed numerical vectors using DistilBERT, which are then incorporated as features to enrich the classical Generalized Linear Model (GLM) framework for insurance pricing. 
Collectively, these works highlight the growing role of NLP in actuarial science, offering innovative approaches to improve predictive accuracy, risk assessment, and decision-making in the insurance industry.

\subsubsection{De-Biasing Features Through Sentiment Analysis} \label{subsubsec:senti}

Some features in insurance datasets are inherently biased due to hidden factors that influence data collection processes, often reflecting underlying societal, demographic, or operational biases. Even the definitions of certain features may implicitly introduce bias, as they can be shaped by subjective interpretations or historical data that fail to capture the full diversity of the population. This bias can skew predictive models, leading to inaccurate risk assessments and potentially unfair outcomes for certain policyholders. NLP can serve as an alternative source to identify, mitigate, and correct biases by extracting unbiased insights from textual data. By analyzing unstructured text, NLP techniques can uncover patterns and nuances that traditional structured data may overlook. For example, sentiment analysis can capture underlying customer satisfaction trends that are not reflected in numerical ratings.

In the insurance domain, customer reviews have emerged as a promising approach to enhance conventional risk quantification, as they often capture critical insights about products and business operations that go beyond conventional rating factors. Thus, customer reviews have the potential to become an emerging rating factor. Our InsurTech dataset contains customer reviews accompanied by a star rating system that numerically represents customer satisfaction. 
Mathematically, the data consists of pairs of textual document (i.e. customer review) and star ratings in the form $\{(d_i,  SR_i)\}_{i=1}^{n}$, where the $i$-th document $d_i=[t_{i,1}, t_{i,2}, \ldots, t_{i,m_i}]$ consists of $m_i$ terms and star ratings $SR_i$ is a integer number from 1 to 5.
These star ratings have initially been incorporated into the InsurTech-enhanced datasets to assess their effectiveness in insurance loss modeling. Surprisingly, both insurers and our InsurTech partner have observed counterintuitive outcomes: Direct use of star ratings as a numerical feature exhibited little to no predictive power in loss modeling, contrary to the prevailing belief that customer reviews should yield predictive improvements. Upon closer examination, we have found that this rating system can introduce biases when used directly in underwriting, as individuals may assign the same star values despite having different attitudes toward a product. For some users, a 5-star rating indicates that the product merely meets their expectations, while for others, a 5-star rating is given only when the product exceeds expectations. This inconsistency raises the question: Can we create a more objective and consistent rating system that relies solely on the text content of customer comments? By applying NLP techniques to analyze the textual data, we can develop a new rating system that better reflects true customer satisfaction, thereby reducing biases and enhancing the reliability of features in insurance applications.

\textit{Sentiment Analysis} is a widely adopted NLP-powered rating system that seeks to quantify the writer's sentiment, specifically the positive or negative orientation toward a given subject. It is commonly viewed as a form of \textit{text classification}, where a label or category is assigned to an entire document.
Thus, the task can be formulated as: given the document $d_i$, we want to predict the sentiment score $SC_i$ of the document as positive or negative. Higher sentiment scores indicate more positive sentiment, whereas lower scores correspond to negative sentiment.
There are two main approaches to sentiment analysis: the Lexicon-Based Approach and the Machine Learning Approach. The Lexicon-Based Approach involves constructing a sentiment lexicon, which can be derived from a specific corpus or built using a general dictionary. In contrast, the Machine Learning Approach can be categorized into supervised and unsupervised methods, depending on data availability. These methods primarily focus on classifying texts into appropriate sentiment categories and predicting sentiment levels.

In our study, we identified a potential bias in the review star rating system, which could compromise the reliability of supervised learning models. Since different individuals may assign different star values despite expressing similar sentiments, using these ratings as labels for training an ML model could introduce bias.
That is, the star ratings $SR_i$ is not an unbiased estimator of the sentiment score $SC_i$, i.e., there exists no such monotone mapping that $SR_i=f(SC_i)$. Instead, the distribution of star ratings may also depend on certain risk characteristics $RC_i$ (e.g., geographical information, industry category), i.e., $SR_i=f(SC_i, RC_i)$. However, in practice, the set of risk characteristics may be data-dependent or difficult to determine, making the bias in star rating systems challenging to mitigate. To confirm this observation, we conduct a Kruskal–Wallis (K–W) test on the distribution of star ratings across states, obtaining a p-value\footnote{To avoid the effect of large sample sizes, we randomly sample 1,000 reviews for 20 times and take the average of the test p-values as the reported p-value.} of 0.0059. The test indicates statistically significant differences in rating distributions among states, suggesting that the direct use of star ratings may introduce bias.
To mitigate this issue, we adopt the \textit{Lexicon-Based Approach} for our purpose. The lexicon-based method typically begins with a seed set of words that carry known sentiment meanings. These seed words are then expanded using large synonym dictionaries such as WordNet\footnote{\href{https://wordnet.princeton.edu/}{https://wordnet.princeton.edu/}}, allowing additional words with similar sentiment features to be incorporated into the lexicon \citep{Whitelaw2005}. Another approach to constructing a sentiment lexicon that is based on statistical and semantic techniques. By analyzing semantic relationships, word similarity, and Pointwise Mutual Information (PMI), it is possible to assign sentiment scores to words in a corpus, thereby building a lexicon with sentiment polarity. Research has shown that using word similarity, lexical association, and, most importantly, distributional similarity, sentiment polarity scores can be systematically assigned to words in a corpus \citep{Read2009}. By applying a lexicon-based sentiment analysis approach, we can generate a more objective sentiment scoring system based on textual content, reducing reliance on the potentially biased star rating system. Given the word, we can retrieve its corresponding sentiment polarity score, which typically ranges between -1 and 1. To calculate the sentiment of a sentence, we average the sentiment score of all words in the sentence using TextBlob\footnote{\href{https://textblob.readthedocs.io/en/dev/}{https://textblob.readthedocs.io/en/dev/}} package.
Following the notation, the polarity score\footnote{For the simplicity of the formulation, we ignore the intensity words such as ``very" and ``not" that modify the polarity score for the sentiment words.}, $PS$, for document $d_i$ is computed as $PS(d_i)=\dfrac{\sum_{t_i\in\mathcal{L}}PS(t_i)}{\sum_{t_i\in\mathcal{L}}1}$ where $\mathcal{L}$ denotes the lexicon dictionary containing sentiment words such as ``good" and ``bad" and polarity score for each sentiment word. By injecting textual sentiment, the polarity score becomes a fairer estimator of sentiment score, i.e., $PS(d_i)=f(SC_i)$. Specifically, the polarity score for each word in our business reviews is derived from SentiWordNet \citep{baccianellaSentiWordNet30Enhanced2010}, a lexical dictionary. Figure \ref{fig:pol} illustrates the decomposition of polarity scores for two 5-star reviews. Words identified by the lexicon-based dictionary are highlighted in red, with their corresponding polarity scores shown below each highlighted term. The overall polarity score for each review is computed as the average of the scores of all identified words, as defined above. The positive expressions such as ``wonderful!" (1.00) and ``great" (0.80) receive high polarity scores, while complaint-related terms such as ``late" (-0.30) and ``terrible" (-1.00) reduce the overall score. This simple aggregation structure allows for straightforward interpretation of sentiment.

\begin{figure}[!ht]
\centering
\begin{minipage}{0.85\textwidth}
\textbf{Example 1:} Everything was handled \wordscore{professionally}{0.10}. The shutters were a \wordscore{few}{-0.20} days \wordscore{late}{-0.30} in coming in; but they were installed \wordscore{recently}{0.00} and look \wordscore{wonderful!}{1.00}\\

\textbf{Example 2:} This place is \wordscore{great}{0.80} for clothing and furniture/home goods; but the book selection is just \wordscore{terrible}{-1.00}. I can't let that count against it; though. I've found some \wordscore{amazing}{0.60} stuff here. I'd recommend checking out the art section. They've got \wordscore{mostly}{0.50} motel art and \wordscore{cheap}{0.40} prints; but I've found a \wordscore{few}{-0.20} \wordscore{really cool}{0.35} \wordscore{amateur}{-0.25} paintings. Oh; and if you need fishing shoes; this is the place for you.
\end{minipage}
\caption{Examples of Polarity Score Decomposition}\label{fig:pol}
\end{figure}

\begin{figure}[!ht]
\centering
\hfill
\subfigure[California vs. Florida]{\includegraphics[scale=0.75]{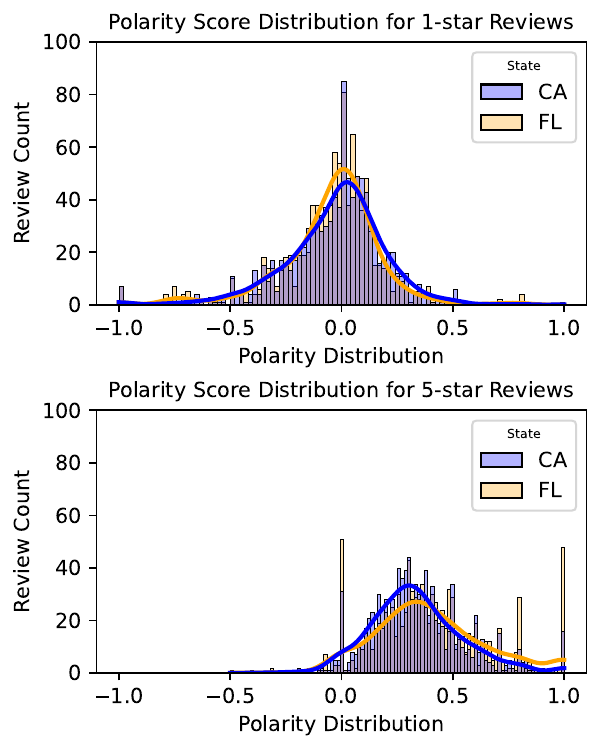}}
\centering
\hfill
\subfigure[California vs. New Jersey]{\includegraphics[scale=0.75]{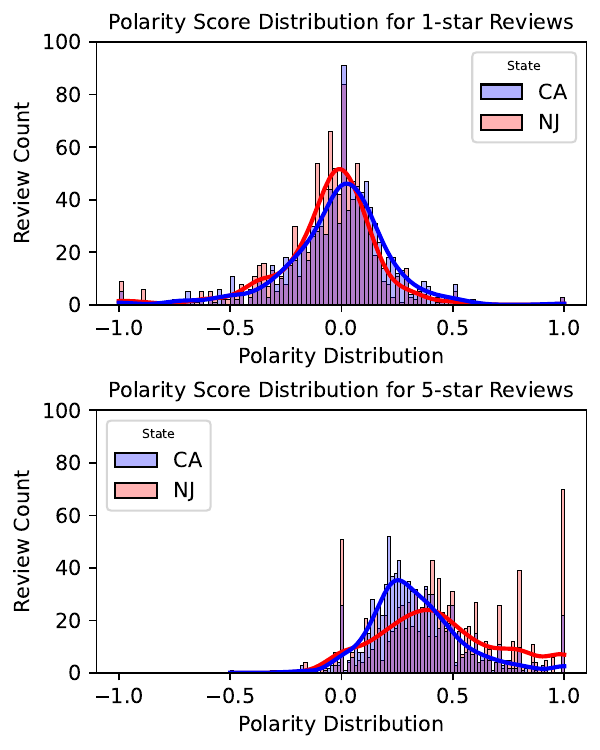}}
\hfill
\caption{Comparison of Review Polarity Distribution by Star Ratings: California vs. Other States}
\label{fig:pol-state}
\end{figure}

We observe that the bias in star ratings extends beyond individual differences and is also evident across geographical regions. Businesses located in different states appear to receive varying treatment in terms of their corresponding star ratings. To explore this phenomenon, we sample 1,000 reviews with extreme star ratings (i.e., 1-star and 5-star) from businesses in California, Florida, and New Jersey. We then compare the distribution of sentiment polarity for these reviews across the three states. As illustrated in Figure \ref{fig:pol-state}, our analysis reveals a striking discrepancy in the polarity distribution of reviews across different geographical regions, even when the star ratings are the same. Notably, for 5-star reviews, businesses in California exhibited a more negative polarity distribution compared to their counterparts in Florida and New Jersey. 
To illustrate this disparity, Table \ref{tab:example-polarity} presents six 5-star reviews from California and Florida, each paired with their respective polarity scores. The Florida reviews exhibit consistently positive sentiment toward the businesses, which aligns closely with the uniformly high polarity scores. In contrast, the California reviews reveal greater variability in sentiment when assessed by human judgment. While some reviews are clearly positive (e.g., Great place excellent service), others are neutral (e.g., I work here) or even negative (e.g., This is bad), despite sharing the same 5-star rating. On the contrary, this variability is accurately reflected by the polarity scores.
This suggests that, despite assigning the highest rating, reviewers in California may have expressed more nuanced or critical sentiments, reflecting less enthusiastic underlying opinions than those from the other states. However, when we examine the polarity distribution of 1-star reviews across the three states, the results are remarkably consistent. The polarity distribution for 1-star reviews shows minimal variation, suggesting that negative sentiment is expressed in a more uniform manner regardless of geographic location. These findings suggest that, while positive sentiment reflected in high star ratings is influenced by regional differences, negative sentiment is more consistent across states. 
To assess whether polarity scores provide a more robust representation than star ratings, we first fit a linear regression model of star ratings on polarity scores and then perform a K–W test on the residuals across states. The test yields a p-value of 0.0409, indicating that the residuals remain significantly associated with geographic information. In contrast, when reversing the regression (i.e., regressing polarity scores on star ratings), the K–W test on the residuals produces a p-value of 0.2271, suggesting that the polarity score has substantially weaker dependence on geographic variation and thus is a fairer representation of sentiment score.
With the rise of neural LMs, a growing body of literature has explored neural LM-based approaches to sentiment analysis \citep{camacho-collados-etal-2022-tweetnlp, barbieri-etal-2022-xlm}, which generally achieve higher classification accuracy. However, most of these models formulate sentiment analysis as a binary classification problem (or a three-class problem when including neutral sentiment), thereby overlooking the intensity of sentiment. As a result, they may be less effective in capturing subtle emotional nuances expressed in reviews. Due to space constraints, additional details on neural LM-based sentiment analysis are provided in Appendix \ref{appnedix_sec:sentiment}.

\begin{table}[!ht]
\small
\centering
\begin{tabular}{cccc}
\toprule
Review Content & State & Star Ratings & Polarity Score \\
\midrule
Great place excellent service & CA & 5 & 0.9 \\
I work here. & CA & 5 & 0.0 \\
This is bad & CA & 5 & -0.70 \\
\hline
THE BEST PLACE TO GET THE BEST VALUE [emoji heart] & FL & 5 & 0.8 \\
Great service and price. [NAME] is awesome. & FL & 5 & 0.9 \\
She's absolutely amazing!!! & FL & 5 & 1.0 \\
\bottomrule
\end{tabular}
\caption{Examples of Review in Star Ratings and Polarity Scores}
\label{tab:example-polarity}
\end{table}

Furthermore, we observe that long reviews often convey mixed sentiments, for example, overall positive impressions of the business coupled with dissatisfaction regarding specific criteria or isolated incidents. 
The two reviews illustrated in Figure \ref{fig:pol} exhibit such mixed sentiments. From a semantic perspective, both reviews are generally positive with respect to the service provided, despite containing minor complaints related to delay or book collection. The polarity score for each sentiment word captures such subtle semantic influence. However, a limitation of the lexicon-based approach is that it assigns equal weight to all words, potentially obscuring the influence of critical terms. For example, in an overall positive review, a disproportionately large number of complaint-related words will lead to a negative polarity score. Similar behavior is also observed in LM–based sentiment analysis methods. In comparison, these nuanced opinions are hidden under star ratings, which fail to accurately reflect the underlying mixed sentiments.
This highlights an inherent bias in star-based rating systems, where cultural, regional, or demographic factors may skew high ratings, leading to potential misinterpretations of customer satisfaction levels. Recognizing this variability is crucial for insurers and businesses alike when incorporating review-based insights into their decision-making processes, as failing to account for these biases could lead to flawed models and misaligned business strategies.

\begin{figure}[!ht]
\centering
\hfill
\subfigure[Accommodation and Food Services vs. Retail Trade]{\includegraphics[scale=0.75]{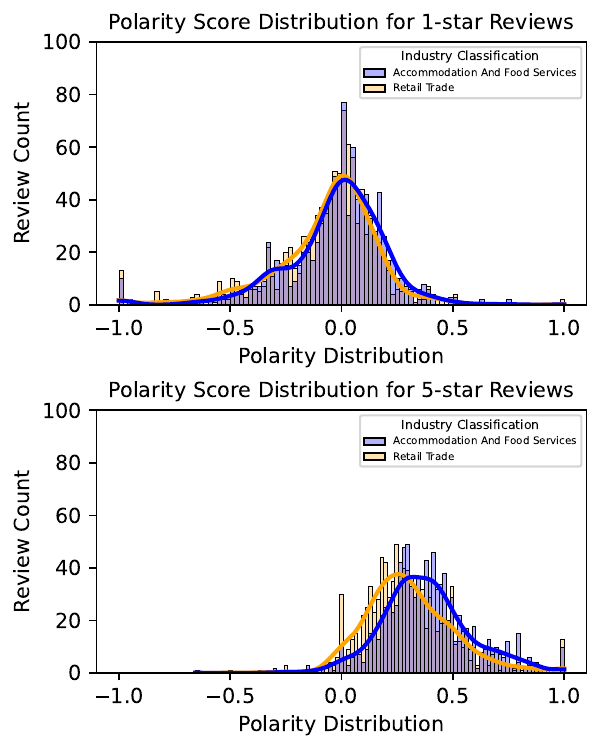}}
\centering
\hfill
\subfigure[Accommodation and Food Services vs. Information]{\includegraphics[scale=0.75]{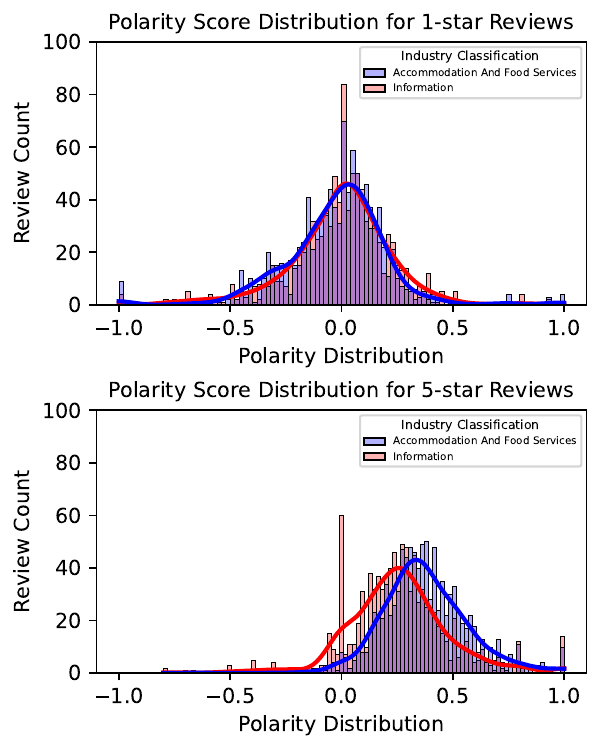}}
\hfill
\caption{Comparison of Review Polarity Distribution by Star Ratings: Accommodation and Food Services vs. Other Industry}
\label{tab:pol-naics}
\end{figure}

Moreover, discrepancies in sentiment polarity distribution are not confined to geographic regions but also extend across different industries. By applying the same process of sampling, sentiment calculation, and plotting as depicted in Figure \ref{tab:pol-naics}, we analyze reviews from three distinct industries: Accommodation and Food Services (primarily consisting of restaurants and hotels), Information Services (including telecommunications, book publishers, and broadcasting centers), and Retail Trade (involving the sale of merchandise for household and personal consumption). Our analysis reveals a similar discrepancy in the sentiment polarity distribution for a given star rating across these industries. Unlike the state-level analysis, where 1-star review polarity shows relatively consistent behavior across different states, the sentiment polarity distribution for 1-star reviews demonstrates significant variability across industries. This inconsistency highlights the complexity of interpreting sentiment from review data and suggests that factors influencing user ratings may differ depending on the industry. The misalignment between star ratings and the underlying reviews may help explain why star ratings exhibit little to no predictive power, further underscoring the need for a more nuanced sentiment analysis approach to account for these variations.
By incorporating polarity score–based sentiment ratings into insurance loss modeling, we find that customer reviews become more informative, addressing previous counterintuitive results and enabling more meaningful model interpretation.

In addition to the qualitative insights provided by the figures, we conducted a quantitative analysis using the Kolmogorov–Smirnov (K-S) test to further validate the differences observed in sentiment polarity distributions. The results, summarized in Table \ref{tab:ks}, confirm our findings by statistically demonstrating significant variations in sentiment polarity across different states and industries. This serves as a compelling illustration of how NLP techniques can enhance fairness by introducing context-aware features derived from textual data, thereby providing a more comprehensive and balanced perspective on the underlying factors influencing outcomes. By incorporating these corrections, insurers can refine their predictive models, mitigate biases, and make more equitable decisions, ultimately ensuring that risk is assessed with greater accuracy and fairness.

\begin{table}[!ht]
\centering
 \begin{tabular}{c c c} 
 \hline
 \textbf{Comparisons} & \textbf{Star} & \textbf{p-value}\\
 \hline
 California vs. Florida & 1-Star & 1.17e-2\\
 California vs. New Jersey & 1-Star &  3.48e-4\\ 
 Accommodation and Food Services vs. Retail Trade & 1-Star & 7.71e-9\\ 
 Accommodation and Food Services vs. Information & 1-Star & 5.36e-1\\  
 \hline
 California vs. Florida & 5-Star & 3.29e-17\\
 California vs. New Jersey & 5-Star&  6.98e-13\\ 
 Accommodation and Food Services vs. Retail Trade & 5-Star & 4.56e-28\\ 
 Accommodation and Food Services vs. Information & 5-Star & 7.79e-43\\  
 \hline
\end{tabular}
\caption{Kolmogorov–Smirnov Test Result of Mentioned Pairs of States and Industries}
\label{tab:ks}
\end{table}

\subsubsection{Transforming High-Cardinality Categorical Features into Numerical Embeddings}

In real-world predictive modeling, it is common to encounter high-cardinality categorical features where the number of distinct categories is large, while the dataset itself is relatively small. This inequality often leads to the curse of dimensionality, a phenomenon that is particularly pronounced in academic settings, where the available data is significantly smaller compared to industry datasets. Traditionally, researchers have addressed this challenge by applying regularization penalties to control model complexity. However, recent advances in NLP offer an alternative approach. \citet{shiNonLifeInsuranceRisk2023} demonstrate the use of categorical embeddings to transform categorical features into numerical vectors within the non-life insurance domain, enabling more effective handling of high-cardinality features. Building on this, \citet{Quan15102024} explore the feasibility of incorporating InsurTech features to enhance business insurance pricing. Specifically, prior work aims to reduce the dimensionality of business categorization by employing a combination of text embedding and clustering techniques.

Our analysis reveals that text embedding and clustering techniques have been proven inadequate for effective feature compression. The InsurTech dataset contains over 500 numerical risk factors, along with raw textual business classifications extracted from various websites. In this work, we retain all numerical features and the response variable, reported claims, while modifying the representation of business classifications. Given our focus on NLP applications in this work, detailed descriptions of the numerical features are omitted and can be found in \citet{Quan15102024}. These raw classifications are highly complex, consisting of 13,287 unique categories with unprocessed and messy text labels, making them difficult to use effectively. In prior work \citep{Quan15102024}, to prevent a drastic increase in dimensionality, these unique business categories have been first embedded into 100-dimensional vectors using word2vec, capturing their static semantic meanings. The embeddings have then been clustered into 24 groups via k-means clustering, aiming to aggregate semantically similar categories. Table \ref{tab:cluster} presents ten such business categories and corresponding category clusters. While this approach has provided a way to encode highly complex categorical variables into more manageable, low-dimensional categories, it introduces two significant drawbacks.
First, some categories, for example, 1- to 5-star hotels, have been grouped into the same cluster (cluster 5) in previous experiments. While these categories share semantic similarities, they may present very different underlying risks. This illustrates that the clustering approach preserves semantic similarity but loses granularity in risk differentiation. Second, static embeddings often fail to capture subtle semantic variations in category titles, introducing information loss. For instance, exchanging the words ``Repair” and ``Service” does not alter the clustering results (cluster 10). However, introducing slight modifications, such as introducing ``Installation” or ``Replacement,” pluralizing ``Repair” to ``Repairs,” or even using ``A/C” instead of ``AC,” can significantly change the resulting cluster assignment. This sensitivity to superficial textual differences is undesirable, as it compromises the robustness and interpretability of the classification. This limitation is not unique to the approaches discussed here; the literature referenced in Section \ref{subsubsec:preact} may also suffer from similar drawbacks. Therefore, instead of collapsing high-dimensional categorical features into lower-dimensional categorical clusters using static embedding, a more effective approach is to transform them into low-dimensional numerical features using context-aware embedding techniques.

\begin{table}[!ht]
\centering
\begin{tabular}{cc}
\toprule
Category Title & Category Cluster \\
\midrule
1 Star(s) Hotel & 5 \\
2 Star(s) Hotel & 5 \\
3 Star(s) Hotel & 5 \\
4 Star(s) Hotel & 5 \\
5 Star(s) Hotel & 5 \\
A/C Installation \& Repair & 19 \\
A/C Repair \& Replacement & 21 \\
AC Repair \& Service & 10 \\
AC Service \& Repair & 10 \\
AC Service \& Repairs & 21 \\
\bottomrule
\end{tabular}
\caption{Examples of Category Titles and Clusters}
\label{tab:cluster}
\end{table}

To further validate the effectiveness of categorical embeddings, we revisit the empirical experiments presented in \citet{Quan15102024}, replacing the 24 cluster-based categorical features with 24 numerical features derived through direct embedding of business categories. Unlike earlier techniques that rely on static semantic representations (e.g., word2vec or GloVe), modern embedding models are trained on vast amounts of textual data, enabling them to capture deeper contextual and semantic nuances. In this study, we adopt the BGE-M3 embedding model introduced by \citet{chenM3EmbeddingMultiLingualityMultiFunctionality2024}, a state-of-the-art multilingual and multifunctional embedding model that produces 1,024-dimensional numerical representations for each business category. As general-purpose LLMs exhibit remarkable transfer learning capabilities, modern NN–based embedding models offer a powerful approach to leverage textual data. Unlike conventional static embeddings, these models generate more effective and informative numerical representations, thereby improving predictive performance in downstream tasks.

\begin{figure}[!ht]
\centering
\includegraphics[scale=0.35]{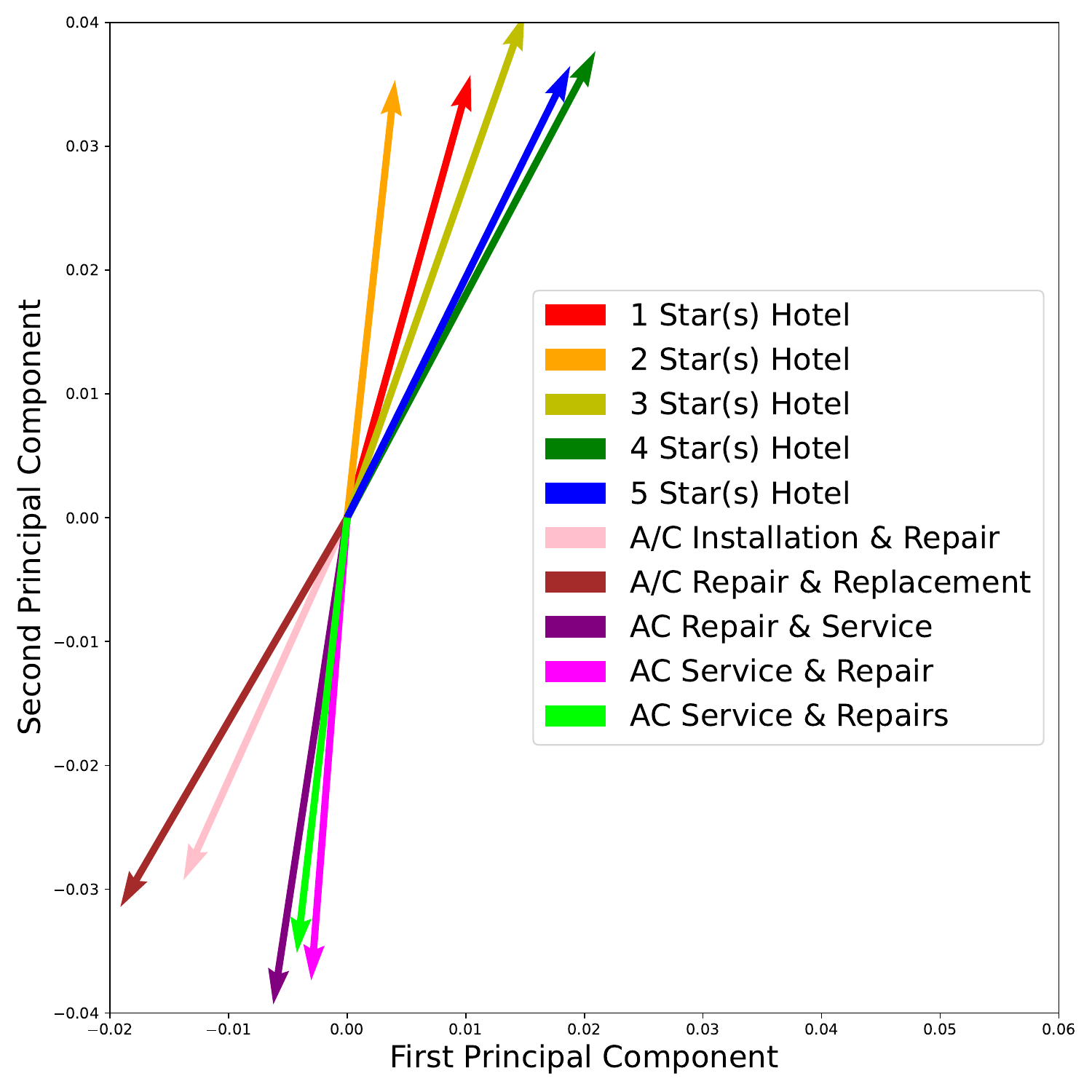}
\caption{Examples of Category Embedding Vectors in First Two Principal Components}
\label{fig:pca-vector}
\end{figure}

However, due to the high dimensionality of these embeddings, direct use in downstream predictive modeling may lead to overfitting or inefficiencies. To address this, we implement the PCA-based dimensionality reduction technique proposed by \citet{raunakEffectiveDimensionalityReduction2019}, selecting the top 24 principal components that explain the greatest variance. This enables us to construct 24 context-aware numerical features that preserve the core semantic information while maintaining compatibility with the experimental setup in \citet{Quan15102024}.
As an illustrative example, Figure \ref{fig:pca-vector} displays the first two principal components of the embeddings for ten category titles (the same as those in Table \ref{tab:cluster}), since the full 24-dimensional vectors are not easily visualizable. It is important to note that because only the first two components are retained and the vector lengths are adjusted for visualization purposes, the figure does not perfectly reflect the actual embeddings used in later empirical experiments. Nevertheless, the plot highlights that category titles with similar semantic meanings yield closer embeddings. For instance, hotel-related titles generally appear in the opposite direction of AC system–related titles. Unlike the static embedding coupled with a clustering approach, the NN-based embeddings capture subtle semantic variations. The lower-star hotels (1- and 2-star) are positioned toward the left of the two-dimensional space, while higher-star hotels (4- and 5-star) shift toward the right. Interestingly, “A/C” and “AC” exhibit greater divergence than expected, whereas word changes such as “Installation,” “Replacement,” and “Service,” which all refer to similar electrical appliance services, produce only minor differences. These results demonstrate that modern neural embedding models, trained on vast textual corpora, can effectively transform high-dimensional categorical variables into practical numerical embeddings while addressing the two key challenges identified in prior approaches.

We then retrain the model with the numerical embedded vectors as numerical features. To evaluate model performance and ensure comparability with prior studies, we adopt four commonly used metrics: Gini index, Percentage Error (PE), Root Mean Squared Error (RMSE), and Mean Absolute Error (MAE) (detailed in Appendix \ref{appendix_sec:eval}). Specifically, the Gini index assesses the discrimination in prediction order, PE captures portfolio predictive uncertainty, and RMSE and MAE quantify individual-level predictive accuracy, with RMSE placing greater emphasis on tail distributions. Together, these metrics provide a comprehensive evaluation of model predictive capabilities. As summarized in Table \ref{tab:VM-result}, 
the insurance in-house model refers to the data provider’s proprietary pricing model, which serves as the baseline. LightGBM with category clustering represents the original approach that applies clustering as a dimensionality reduction technique, both reported in the \textit{BG} section of Table 4 in \citet{Quan15102024}. By contrast, LightGBM with category embedding introduces the new numerical embedding method, where high-dimensional categorical features are transformed into low-dimensional, context-aware numerical embeddings, leading to improved model performance.
The Gini index and PE show no significant improvement, while RMSE and MAE exhibit modest reductions. This suggests that replacing category clustering with category embeddings does not substantially alter the overall model structure. Instead, it primarily affects local behavior within the model, leading to adjustments at specific leaf nodes in the LightGBM model.
It should be noted that the BG coverage includes 561 features, of which the 24 category clustering/embedding features constitute only a small fraction of the overall feature space. The rest of the numerical 537 features, such as additional risk characteristics and industry indicators, also contribute to interpreting the business category, thereby reducing the relative influence of our engineered features. Consequently, the performance gain of category embedding is expected to be moderate at best. Moreover, the complexity of real-world InsurTech datasets further limits the potential for substantial metric improvements. Although we retain only 24 numerical features for consistency, further increasing the number of retained components may yield additional gains, leaving space for further significant improvement. These findings reinforce the value of NLP-driven categorical embedding techniques for transforming high-cardinality categorical variables into compact, informative numerical features, ultimately improving predictive modeling in insurance by better capturing underlying semantic structures.

\begin{table}[!ht]
\centering
\begin{tabular}{llrrrr}
\toprule
Split & Model & Gini & PE & RMSE & MAE \\
\midrule
\multirow{3}{*}{train} & Insurance in-house model & 0.29 & -0.40 & 5761.94 & 1526.47 \\
& LightGBM + category clustering & 0.84 & \textbf{0.00} & 5364.05 & 1198.07 \\
& LightGBM + category embedding & \textbf{0.85} & \textbf{0.00} & \textbf{5347.50} & \textbf{1196.14} \\
\hline
\multirow{3}{*}{test} & Insurance in-house model & 0.32 & -0.54 & 5328.02 & 1461.92 \\
& LightGBM + category clustering & 0.37 & \textbf{-0.08} & 5198.57 & 1181.47 \\
& LightGBM + category embedding & \textbf{0.38} & \textbf{-0.08} & \textbf{5162.89} & \textbf{1177.69} \\
\bottomrule
\end{tabular}
\caption{Model Performance based on Validation Measures}
\label{tab:VM-result}
\end{table}

\begin{figure}[!ht]
\centering
\hfill
\subfigure[Double Lift Chart for LightGBM Models]{\includegraphics[scale=0.45]{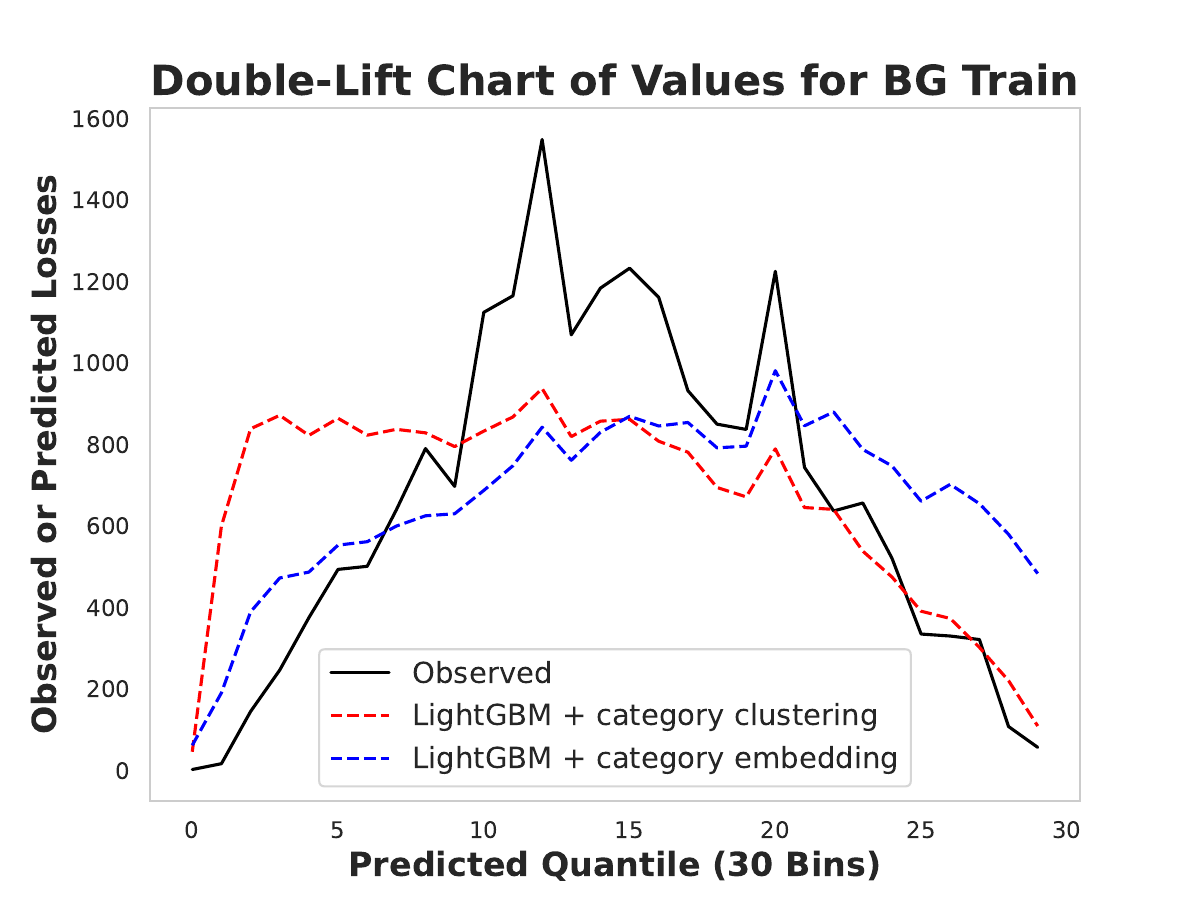}\label{fig:dl_bg}}
\centering
\hfill
\subfigure[Top 25 Gini feature importance of LightGBM + Category Embedding]{\includegraphics[scale=0.4]{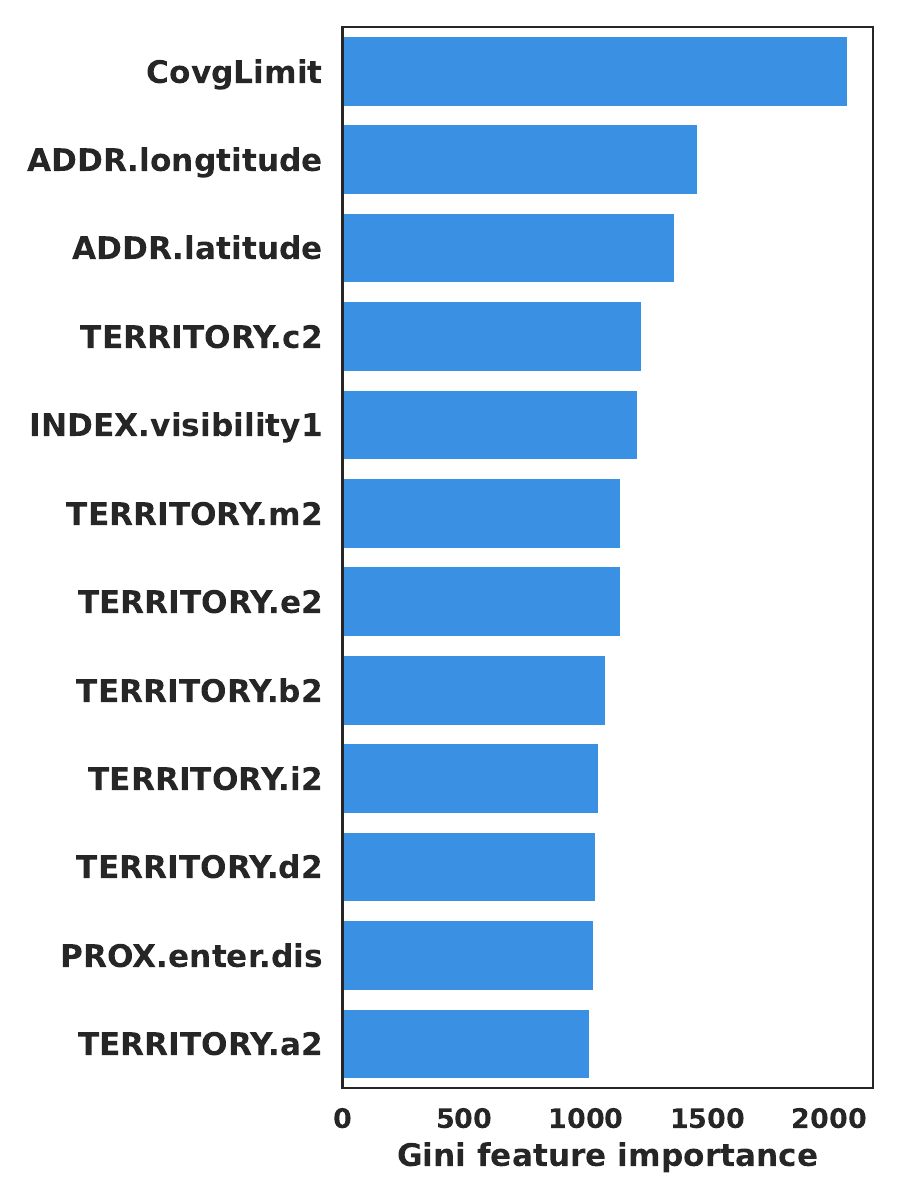}\label{fig:gini_bg}}
\hfill
\caption{Interpretability of Numerical Embeddings}
\label{fig:embedding}
\end{figure}

Given the regulatory emphasis on model interpretability in the insurance sector, we examine how numerical embeddings affect model performance. Figure \ref{fig:dl_bg} presents a double-lift chart comparing two LightGBM models: one using category clustering and the other using learned embeddings. In contrast to \citet{Quan15102024}, where the comparison is made between an in-house insurance model and an InsurTech-enhanced LightGBM model (category clustering), the two models considered here exhibit more similar behavior, as they are trained on largely overlapping datasets. 
Under the construction of the double-lift chart, the predicted quantiles (x-axis), ordered from left to right, represent underestimation to overestimation in the predictions of the category embedding model relative to the category clustering model. The results indicate that the embedding-based model better captures claim losses in the lower quantiles (i.e., smaller losses), while exhibiting a slight degradation in the upper quantiles. This pattern suggests that smaller claim losses benefit from more granular feature representations, which enable the model to capture subtle differences in underlying risk profiles. In contrast, less granular representations, such as category clustering, appear to be more effective in capturing extreme claim losses. However, this enhanced differentiation has a limited impact on global feature importance. As shown in Figure \ref{fig:gini_bg}, the top 25 features remain largely consistent with those identified by the category clustering model in \citet{Quan15102024}. The category clustering/embedding features, denoted as \textit{CLASS.catX} (where X ranges from 0 to 23), do not appear among the top 25 features in terms of importance in either the clustering-based or embedding-based models.

\subsection{Standalone NLP Use Cases}

We have demonstrated in Section \ref{subsec:data} how NLP can be used for data enrichment, where textual insights augment structured datasets to support broader predictive modeling objectives. In such scenarios, NLP techniques typically serve as subtasks or auxiliary components that enhance the performance of primary actuarial models. In this section, we shift our focus to standalone NLP use cases in actuarial science—applications where NLP methods deliver direct value.

In the actuarial science literature, there are relatively few standalone NLP use cases to date. For instance, \citet{Liao2020} apply traditional text mining and classification techniques to customer service call transcripts, enabling insurers to categorize and process calls more efficiently—ultimately saving time and resources. \citet{Zhang11112024} develop an NLP-powered repository and search engine focused on cyber risk literature, culminating in a web-based tool that facilitates access to relevant research in this emerging area. Additionally, \citet{cao2024assessing} leverage legal documents to build a claim dispute prediction model using NLP, offering a novel approach to quantifying insurers' litigation risk. \citet{liTextualAnalysisInsurance2025} propose an LLM-powered framework to assist insurance claims processing by classifying insurance claims using distance-based metrics and analyzing semantic relationships through prompt-based methods.
Although limited in number, these studies demonstrate the versatility and practical value of NLP in insurance applications, suggesting ample opportunities for future exploration. To stimulate discussion and inspire future development, we present several representative standalone use cases of NLP in actuarial and insurance contexts. Given the rapid advancement in LLMs, this list is by no means exhaustive and is expected to grow over time. 

\textbf{Text and Speech Processing}: These include foundational methods such as tokenization (introduced earlier), as well as more advanced applications like Optical Character Recognition (OCR) and Speech Recognition. OCR enables insurers to digitize printed or handwritten documents—particularly valuable for converting legacy records into analyzable text. Similarly, speech recognition transforms audio recordings (e.g., customer service calls) into text, allowing for NLP applications such as sentiment analysis or conversation mining. These technologies are essential for insurers, who are often legally required to maintain archival records for decades, much of which exist in non-digital formats.

\textbf{Lexical Semantics}: This category includes tasks such as sentiment analysis (discussed previously) and Named Entity Recognition (NER), which identifies proper nouns and classifies them into categories such as persons, locations, or organizations. In the life insurance domain, timely identification of deceased policyholders is crucial. However, in jurisdictions like the United States, where centralized death records are unavailable or fragmented, this can be challenging. NLP systems equipped with NER can automatically monitor obituary websites and public notices to detect policyholder death events in near real-time, offering a powerful tool for mitigating annuity fraud and overpayments.

\textbf{Text Summarization and Conversational AI}: NLP can also be applied to generate concise summaries from lengthy text documents, which is especially useful for parsing dense insurance policy language or legal documents. Additionally, the emergence of question-answering systems and AI chatbots allows insurers to provide interactive, automated support for customer inquiries by retrieving relevant information from large document repositories or past interactions.

While these examples highlight a broad range of standalone NLP use cases, we further illustrate one in detail: industry classification based on business descriptions. This use case integrates topic modeling (a form of text summarization) and text classification to automatically infer a company's industry sector from its text descriptions. We elaborate on the methodology and implementation of these techniques in the following subsection.

\subsubsection{Business Industry Classification}\label{subsubsec:class}

One of the most critical rating factors in commercial insurance pricing is industry classification, i.e., the industry to which a business belongs. 
This classification serves as the foundation for underwriting and risk assessment, as risk exposure varies significantly across industries. While the process might appear straightforward, accurately classifying a business's industry is far more complex than it seems. Currently, insurers primarily rely on self-reported information from business owners \citep{dumbacherBEACONToolIndustry2025}, combined with underwriters' professional judgment, to determine industry classifications. However, this approach often lacks consistency and structure, leading to potential misclassifications. Such subjectivity not only introduces pricing unfairness but also exposes insurers to unforeseen risks due to incorrect industry classification. Adding to the challenge is the fact that many modern businesses span multiple industries. For example, consider a brewery: it manufactures alcoholic beverages (suggesting a manufacturing classification), sells bottled products directly to consumers (fitting a retail profile), and may also operate an on-site bar or restaurant (aligning with food services). Determining a single industry classification for such a multifaceted operation is not only difficult for business owners but also confusing for underwriters. Misclassification in these cases can lead to inappropriate pricing, coverage gaps, or even regulatory issues. As businesses become increasingly hybrid and diversified, the need for a more systematic and data-driven approach to industry classification becomes crucial to ensuring both actuarial fairness and underwriting accuracy.

\begin{figure}
\centering
\includegraphics[scale=0.4]{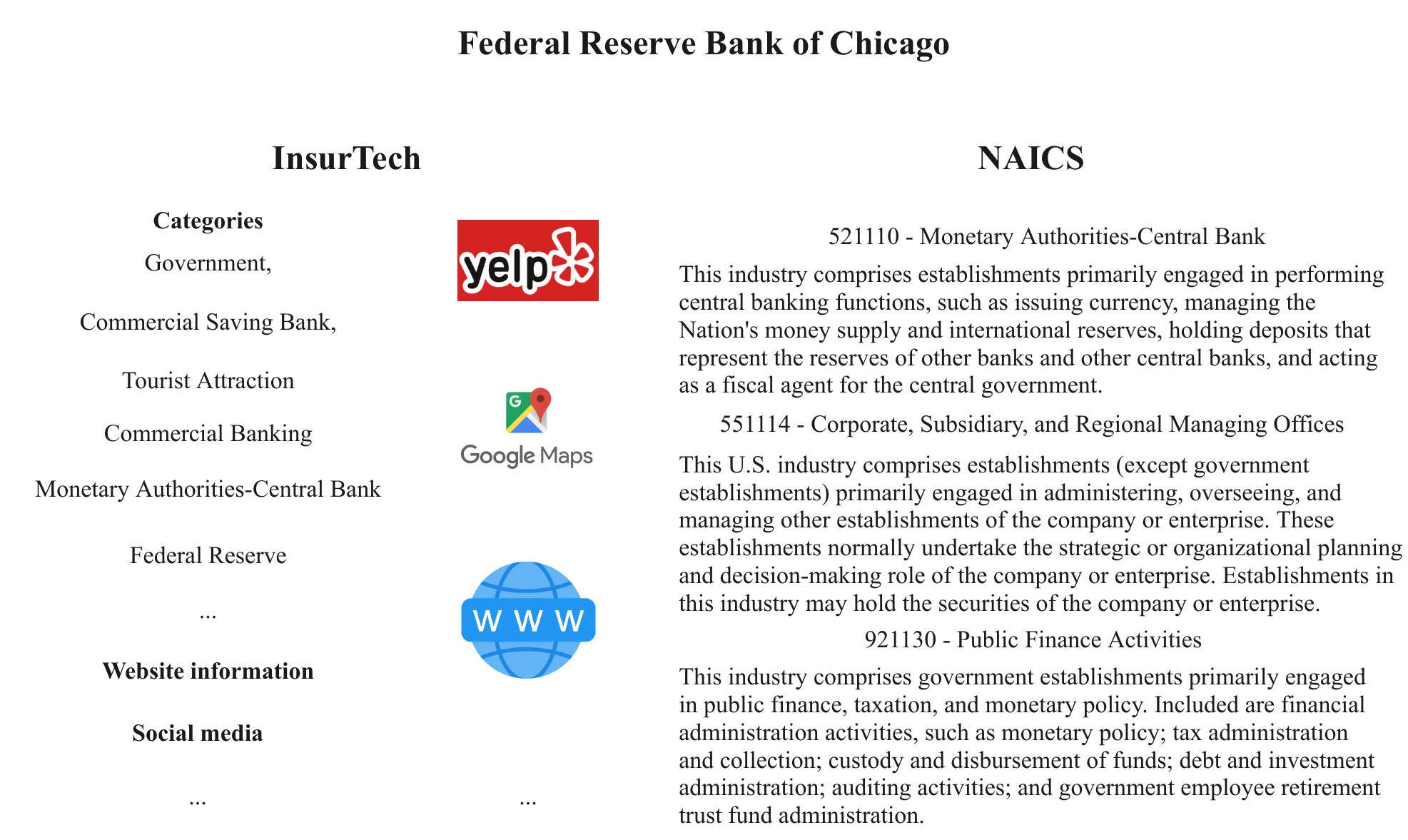}
\caption{Summary and Descriptive Data for Federal Reserve Bank of Chicago}
\label{fig:data-source}
\end{figure}

One widely used industry classification framework is the North American Industry Classification System (NAICS)\footnote{https://www.naics.com/}, the standard used by U.S. federal statistical agencies to classify business establishments for purposes such as data collection, analysis, and reporting.
Widely adopted across industries and government, NAICS currently encompasses 17,768,531 U.S. business entities\footnote{https://www.naics.com/search/}. Its hierarchical structure is defined by six digits: the first two identify broad industry sectors, while additional digits provide increasingly granular classification. Specifically, the three- to six-digit levels represent subsector, industry group, NAICS industry, and national industry, respectively. In this way, each digit encodes a more detailed layer of industry hierarchy.
For example, a commercial bank ($522110$, e.g., JPMorgan Chase Bank) and a credit union ($522130$, e.g., Navy Federal Credit Union) cannot be distinguished at the two-digit sector level ($52$: Finance and Insurance) or even at the four-digit industry group level ($5221$: Depository Credit Intermediation). By contrast, a central bank ($521110$, e.g., the Federal Reserve Bank of Chicago, shown in Figure \ref{fig:data-source}) can be differentiated from commercial banks or credit unions at a more generalized level (four-digit). This illustrates that finer NAICS codes capture a more precise industry classification. However, such granularity comes with significant complexity. In the most recent 2022 NAICS version\footnote{https://www.census.gov/naics/2022NAICS/2-6\%20digit\_2022\_Codes.xlsx}, there are only 20 unique two-digit sectors, but 308 four-digit industry groups and 1,012 six-digit national industries. This sharp increase in class complexity highlights a trade-off between classification granularity and practical accuracy, posing challenges both for insurers and for broader applications such as government economic data collection (U.S. Census Bureau) and tax collection for the Internal Revenue Service (IRS). Thus, it is critical for insurers to carefully determine the level of NAICS granularity appropriate for their analytical and operational needs. Moreover, NAICS codes also face the issue of multi-classification for a single business. For instance, Home Depot is officially classified as both $444120$ (Paint and Wallpaper Retailers) and $444180$ (Other Building Material Dealers). Such overlaps complicate industry classification and risk assessment.

InsurTech, through its dynamic data collection processes, provides NLP solutions to enhance the accuracy and granularity of NAICS-based industry classification. Specifically, we leverage two key sources of information, InsurTech data and official NAICS definition information, to perform industry classification. InsurTech aggregates business information from a wide range of online sources, including business categories scraped from platforms like Yelp, Google Maps, and other online directories, website information, and social media. In contrast, the official NAICS definition information provides accurate code titles and detailed descriptions. For instance, Figure \ref{fig:data-source} illustrates the data utilized to classify the Federal Reserve Bank of Chicago. InsurTech data summarizes categories across multiple platforms, yielding short phrases such as Government and Monetary Authorities—Central Bank. We may assign the institution with three potential NAICS codes: $521110$ (Monetary Authorities—Central Bank), $551114$ (Corporate, Subsidiary, and Regional Managing Offices), and $921130$ (Public Finance Activities), each accompanied by precise definitions specifying their scope.
While humans can read, interpret, and synthesize information from both InsurTech data and official NAICS definition information to make a reasoned classification decision, 
professionals or even experts can fail on large-scale classification tasks, especially for the six-digit level with 1,012 classifications. Thus,
our goal is to automate this process. We develop an NLP-driven framework that replicates this human judgment, using advanced language models to bridge business information with structured industry classifications and assign the appropriate NAICS codes.

From an NLP technical perspective, there are two main approaches to building an automated business industry classification system. The first approach is to frame the problem as a \textit{text classification} task, where given a piece of text, the model predicts the group or class it belongs to. In fact, sentiment analysis, which we discussed earlier in Section \ref{subsubsec:senti}, is a specific form of text classification. Even language modeling mentioned in Section \ref{subsec:embedding} can be viewed through this lens: predicting the next word is essentially classifying the current context into one of many possible word classes. A typical text classification pipeline involves three major steps: feature extraction, dimensionality reduction (if necessary), and classification. However, this approach requires a well-labeled dataset where each piece of text is clearly assigned to a target class. For example, the BEACON system, an industry classification tool proposed by \citet{dumbacherBEACONToolIndustry2025}, is trained on proprietary U.S. Census data containing 4.3 million observations, allowing it to adopt a supervised learning approach. In our case, the dataset contains information for $13,873$ businesses. Among them, only $2,372$ of these businesses have confirmed NAICS codes and proper descriptive information available on their official webpages. If we were to pursue a supervised text classification approach, we would have to discard the majority of our data due to missing labels or non-informative features, significantly limiting the amount of training data available. Furthermore, the vast number of categories at the four- and six-digit levels results in severe underrepresentation for most of the codes, raising concerns about the reliability and credibility of the resulting model.

Therefore, we adopt the second approach: unsupervised learning leveraging \textit{topic modeling}. Topic modeling aims to uncover latent thematic structures within large collections of text by analyzing the co-occurrence patterns of words across documents. Through this process, it identifies coherent sets of words that together represent distinct ``topics." In our study, we employ well-known and interpretable topic modeling techniques: Latent Dirichlet Allocation (LDA) \citep{Blei2003} and Rapid Automatic Keyword Extraction (RAKE) \citep{Stuart2010}. LDA is a generative probabilistic model designed to represent documents as mixtures of topics, where each topic is characterized by a distribution over words. Specifically, LDA assumes that each document $d_j$, $j=1, \dots, n$ is a mixture of a finite number of topics $z_k$, $k=1\dots Z$, and each topic is characterized by a distribution over words. The main parameters of LDA include $\theta_j$ and $\phi_k$. $\theta_j$ is the topic distribution for document $d_j$, $j=1, \dots, n$ and it is drawn from a Dirichlet distribution with parameter $\alpha$, i.e., $\theta_j \sim \text{Dirichlet}(\alpha)$. $\phi_k$ represents the word distribution for topic $z_k$, $k=1\dots Z$, drawn from a Dirichlet distribution with parameter $\beta$, i.e., $\phi_k \sim \text{Dirichlet}(\beta)$. For each word $t_i$, $i=1, \dots, m$ in the document $d_j$, $j=1, \dots, n$, a topic $z_{d_j,t_i}$ is chosen from a multinomial distribution $\text{Multinomial}(\theta_j)$, and a word $t_i$ is selected based on $\text{Multinomial}(\phi_{z_{d_j,t_i}})$. Mathematically, the joint distribution over topics, words, and documents in LDA is given by:
\begin{equation}
    P(\textbf{t}, \textbf{z}, \theta, \phi, |\alpha, \beta) = \prod_{k=1}^Z P(\phi_k|\beta) \prod_{j=1}^n P(\theta_j|\alpha) \prod_{i=1}^m P(z_{d_j,t_i}|\theta_j)P(t_i|\phi_{z_{d_j,t_i}})
\end{equation}
This objective function closely resembles those discussed in Section \ref{subsec:embedding} in the context of LM. However, topic modeling introduces additional assumptions by explicitly modeling latent topics through Bayesian inference. Specifically, LDA estimates these hidden topic structures, enabling the extraction of coherent word groupings and deeper insights from large-scale textual data. Moreover, the resulting topic distributions provide a principled basis for measuring document similarity based on shared thematic content, adding an interpretable layer for downstream analytical tasks. RAKE is another widely used algorithm for extracting keywords from individual documents. Its core idea is to identify keywords by first removing non-informative words, such as stopwords. Once these are stripped, RAKE generates candidate keywords and ranks them using word frequency, word degree (co-occurrence with other words), or their ratio. These metrics help surface the most relevant keywords without relying on complex linguistic analysis. For brevity, we omit the algorithm's mathematical formulation.

\begin{figure}[!ht]
\centering
\includegraphics[scale=0.48]{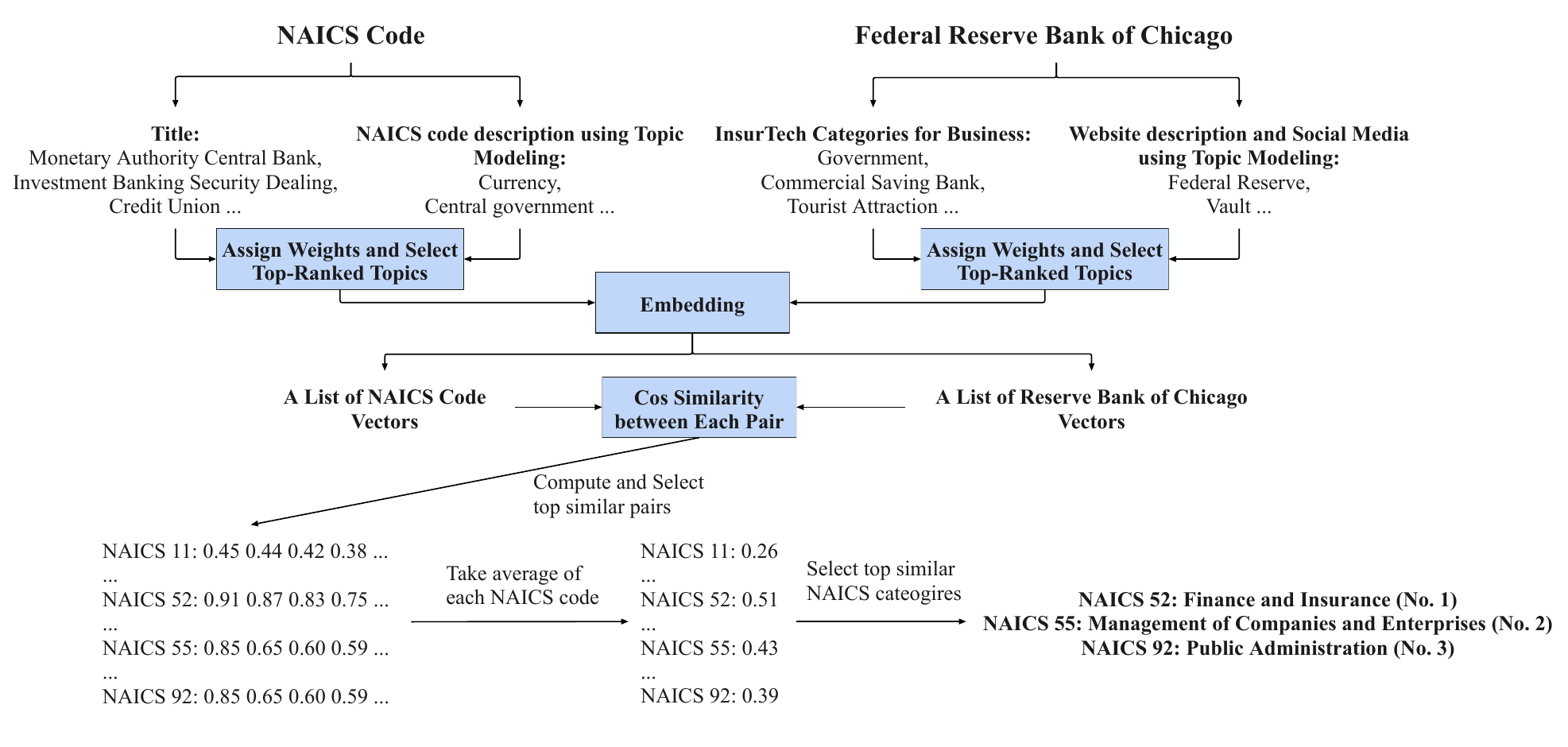}
\caption{Unsupervised Industry Classification Pipeline}
\label{fig:unsup-pipe}
\end{figure}

\begin{figure}[!ht]
\centering
\includegraphics[scale=0.65]{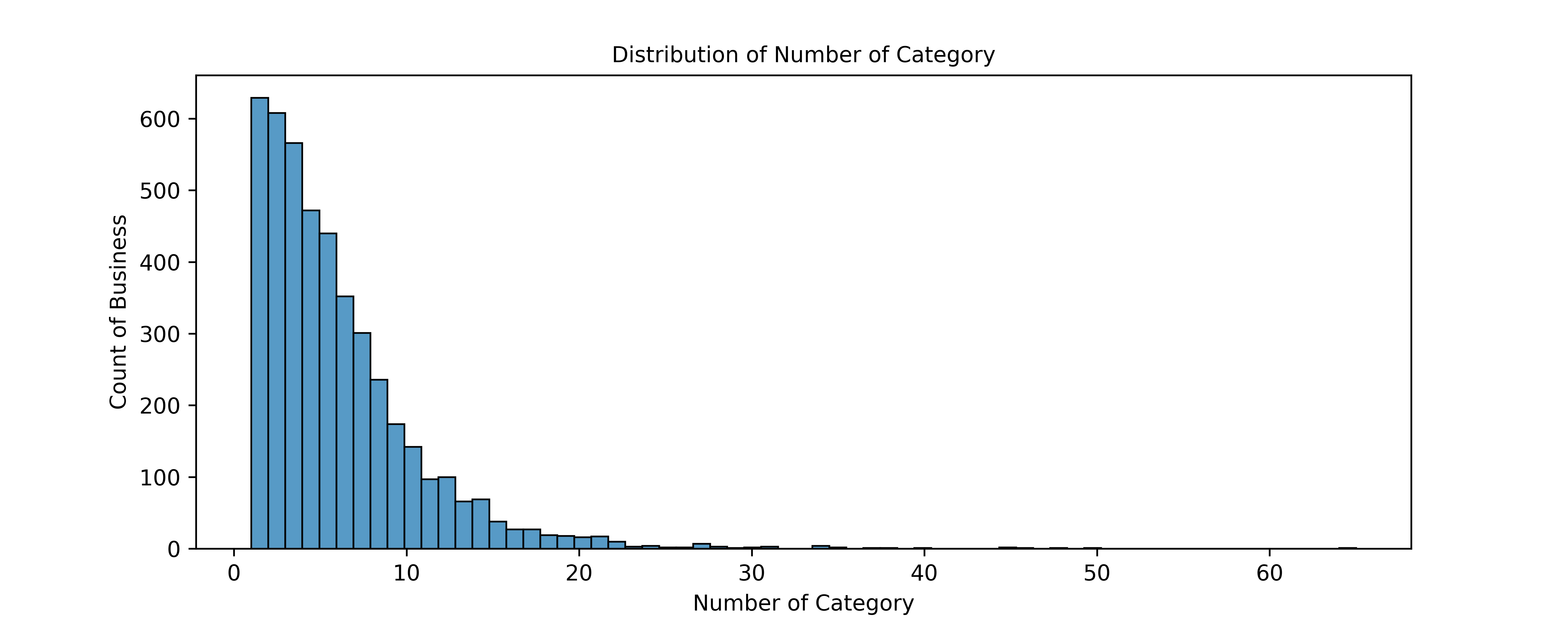}
\caption{Distribution of Number of InsurTech Categories for Each Business}
\label{fig:unsup-cate}
\end{figure}

Specifically, we apply LDA to business reviews and use RAKE to extract keywords from business website descriptions. 
To demonstrate our unsupervised learning approach, we revisit the Federal Reserve Bank of Chicago example, whose classification pipeline is summarized in Figure \ref{fig:unsup-pipe}. From the official NAICS definition information, each NAICS code is represented by its titles (e.g., Monetary Authority Central Bank) and the topic words we extracted from NAICS code descriptions (e.g., Currency, Central government).
In parallel, the business entity, the Federal Reserve Bank of Chicago, is described using categories engineered by InsurTech (e.g., Government, Commercial Saving Bank) and topic words extracted from business website descriptions and social media, like Yelp and Google Maps (e.g., Federal Reserve, Vault). While InsurTech categories provide concise summaries of business activities, their availability varies, with some businesses have few or none. For instance, the Federal Reserve Bank of Chicago has only five such categories. Figure \ref{fig:unsup-cate} shows that the distribution of available InsurTech categories is highly right-skewed, motivating us to supplement missing or sparse InsurTech categories using topic modeling from business website descriptions and social media.
To identify the most relevant topic words for downstream tasks, we further filter these topic words by measuring their semantic similarity to existing InsurTech categories. Top-matching words are selected to enrich or replace missing InsurTech categories. Similarly, for businesses with many InsurTech categories, we can filter and retain only those that are most relevant based on similarity to topic words. This approach also applies to NAICS code titles and descriptions, though titles alone often suffice.

Using the embedding mentioned in Section \ref{subsec:embedding}, we represent each business as vectors based on its InsurTech categories and extracted topic words. A similar embedding process is applied to the NAICS code descriptions. While various embedding choices are available, we omit detailed comparisons here and refer interested readers to \citet{cao2024assessing} for further discussion. We select the highest-dimensional version of the chosen embedding model, as higher-dimensional vectors typically capture more semantic information and yield better performance in downstream tasks. Specifically, we adopt the MiniLM embedding model \citep{wangMINILMDeepSelfattention2020} provided by \textit{sentence\_transformer}\footnote{https://sbert.net/\#}.

We then reformulate the most probable industry classification task as a business–NAICS similarity problem. Specifically, we compute the pairwise similarity between business and NAICS embedding vectors using \textit{Cosine Similarity}, which measures the angle between two vectors and yields a score ranging from $-1$ to $1$, indicating how closely aligned the business and NAICS code representations are in the vector space.
\begin{equation}
    Similarity(v_{t_{i}}, v_{t_{j}}) = \frac{v_{t_{i}}\cdot v_{t_{j}}}{|v_{t_{i}}|*|v_{t_{j}}|}
\end{equation}
where $v_{t_{.}}$ is the embedding for the word. 

Once similarity scores are computed between all business–NAICS embedding pairs, we retain the top \textit{P} most similar pairs to mitigate the influence of noisy textual embeddings on the score distribution, and average these scores to obtain the overall similarity between the business and the given industry. Repeating this process across all NAICS codes allows us to rank industries by relevance for the business. 
For the Federal Reserve Bank of Chicago, the top three 2-digit NAICS codes with the highest similarity scores are: Code 52 (\textit{Finance and Insurance}, similarity of 0.51), Code 55 (\textit{Management of Companies and Enterprises}, similarity of 0.43), and Code 92 (\textit{Public Administration}, similarity of 0.39). These codes represent the most probable sectors identified by our unsupervised learning pipeline. Validation against the true NAICS classification confirms that two of these codes, 52 and 55, align with the organization's recorded categorization in our dataset. While underwriters would officially classify the Federal Reserve Bank of Chicago under the Finance industry, its role as the central bank of the United States also makes a reasonable case for inclusion under the Public Administration category. By accurately classifying the Federal Reserve Bank of Chicago under the Finance industry, insurance underwriting systems can then differentiate the business from other sectors (e.g., construction, manufacturing), which inherently carries different operational risks. In addition to accurate industry code classification, our unsupervised learning approach can also function as a decision-support system through its output of category similarity scores. In practice, underwriters can review the most probable categories alongside their associated probabilities and make informed decisions when the system demonstrates high confidence. Conversely, when the probabilities are low or when the suggested categories appear questionable, the system can serve as a trigger for further investigation of the business information. This human-in-the-loop framework reduces the overall overhead of industry classification while maintaining high accuracy and reliability.

\begin{figure}[h!]
\centering
\subfigure[2-digit Unsupervised Industry Classification]{\includegraphics[scale=0.4]{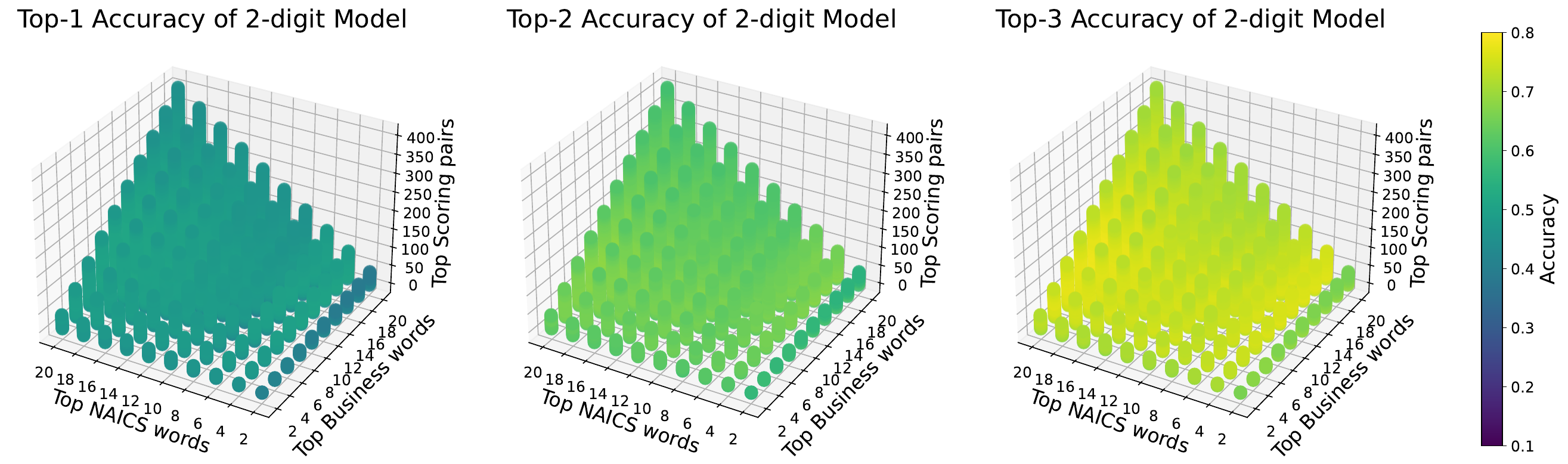}\label{fig:2-digit}}
\subfigure[4-digit Unsupervised Industry Classification]{\includegraphics[scale=0.4]{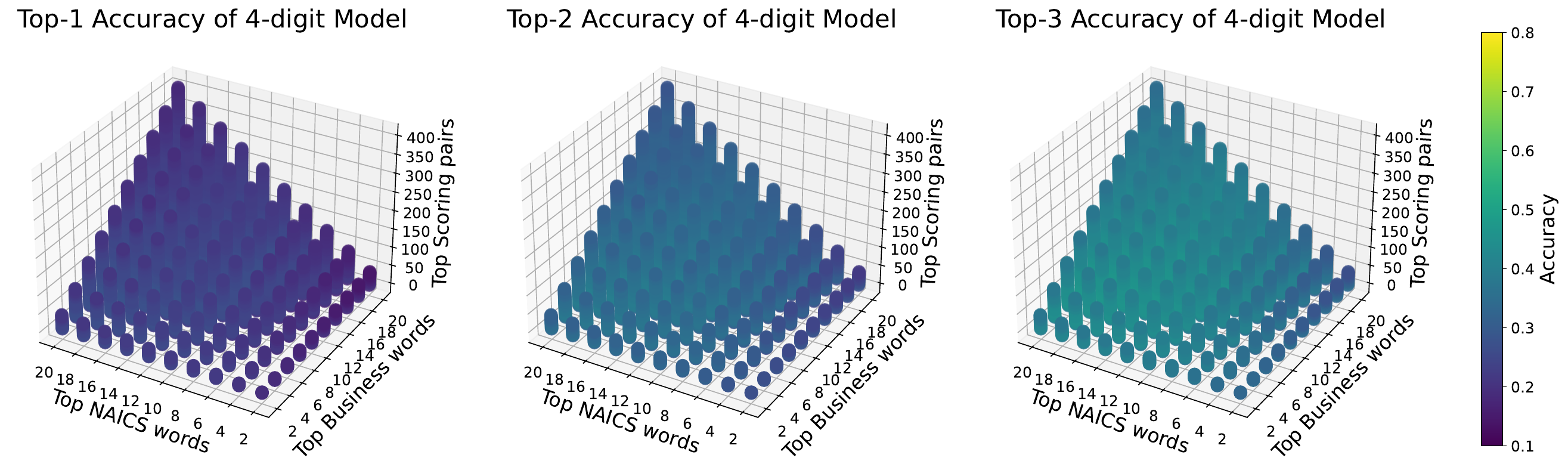}\label{fig:4-digit}}
\subfigure[6-digit Unsupervised Industry Classification]{\includegraphics[scale=0.4]{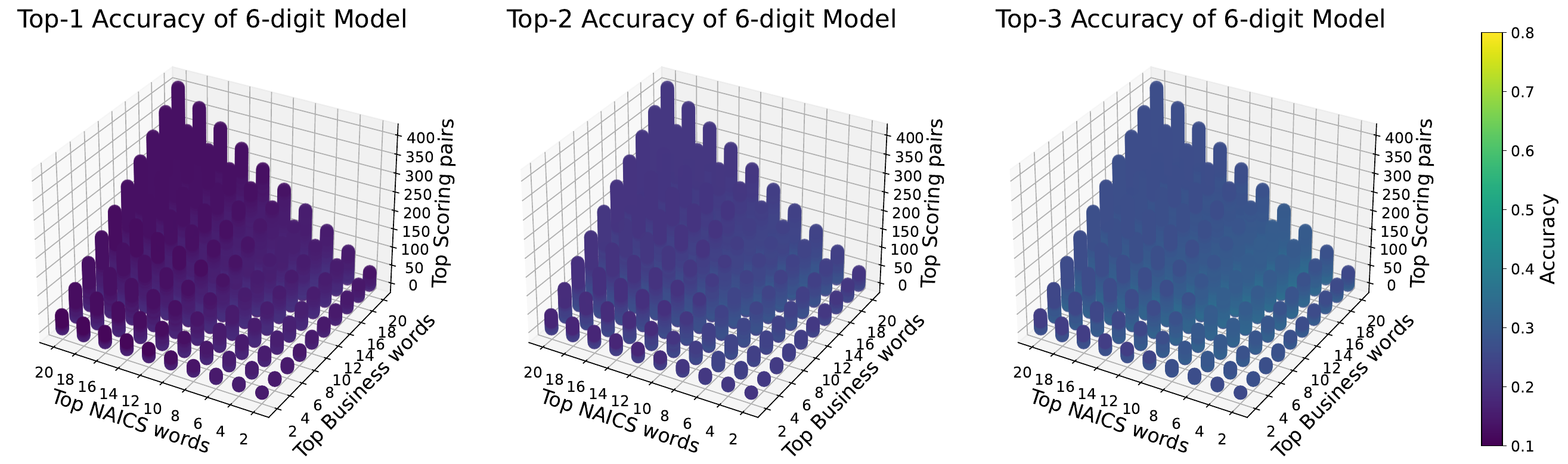}\label{fig:6-digit}}
\caption{Impact of Hyperparameters on Unsupervised Industry Classification}
\label{fig:unsup-hyper}
\end{figure}

This unsupervised strategy avoids dependence on limited or potentially inaccurate NAICS labels. Still, for evaluation, we compare our results against these weak labels to assess performance. As discussed above, our unsupervised learning solution includes three key tunable hyperparameters regarding top topics/similarity pairs selection: (1) the number of topic words used to represent each NAICS code, (2) the number of topic words used to represent each business, and (3) the number of top-scoring word pair, $\lambda$, embeddings averaged per comparison. 
Each NAICS code and business description contains a mix of topic words—some highly informative for unsupervised classification, others serving as noise. In addition, certain sources may include redundant terms, which can reduce the efficiency of similarity computation and classification. To address this, the first two hyperparameters regulate both the quantity and quality of topic words extracted from each NAICS code and business description. Pairwise similarity scores between topic-word embeddings are then computed, and the top $\lambda$ matches ($\lambda\leq\text{number of NAICS topic words}\times\text{number of business topic words}$), determined by the third hyperparameter, are used to categorize the business industry. 
Adjusting these hyperparameters improves topic representation and reduces noise from irrelevant word pairs. In our experiment, we vary all three hyperparameters to evaluate their impact on top-1, top-2, and top-3 predictive accuracy, where the top-k predictions are used to assess each model's classification performance.
Here, top-k prediction accuracy counts if there is a match among the top k most probable candidates to the true answer, which is widely used in the industry. The optimal hyperparameter configuration, determined based on top-1 accuracy, is summarized in Table \ref{tab:unsup-hyper}. The corresponding evaluation metrics for this configuration are presented in Table \ref{tab:unsup-result}, with detailed definitions provided in Appendix \ref{appendix_sec:eval}. In addition, Figure \ref{fig:unsup-hyper} visualizes the effect of the three hyperparameters on classification accuracy. Each plot depicts the top-1, top-2, or top-3 classification accuracy for the 2-, 4-, or 6-digit NAICS code prediction task. The three axes correspond to the three hyperparameters under evaluation. A unified color gradient is employed across all hyperparameter settings to visualize their impact: brighter hues (ranging from green to yellow) indicate higher model accuracy, while darker shades (from blue to purple) represent lower accuracy levels. 
Allowing more NAICS or business-related keywords to contribute to similarity scoring and retaining more scoring pairs for the final prediction generally improves the model's ability to classify businesses accurately. This underscores the valuable role that InsurTech-driven categorization can play in improving industry classification. However, as the classification becomes more granular—moving from broader 2-digit codes to more specific 4- or 6-digit codes—the model's classification performance declines 
(darker shades are observed on the lower plots)
. This decrease is likely due to a substantial increase in category complexity, combined with the limited availability of high-quality descriptive textual data. Nevertheless, the significant improvements in top-2 and top-3 prediction accuracy indicate that the correct classifications often appear among the highest-ranked results.

\begin{table}[!ht]
\centering
\begin{tabular}{cccc}
\toprule
Digits & \multicolumn{3}{c}{Optimal Hyperparameter Sets} \\
& NAICS words & business words & scoring $\lambda$ pairs \\
\midrule
2 & 20 & 16 & 84 \\
4 & 20 & 20 & 24 \\
6 & 12 & 18 & 14 \\
\bottomrule
\end{tabular}
\caption{Optimal Hyperparameter Setting of Unsupervised Industry Classification}
\label{tab:unsup-hyper}
\end{table}

\begin{table}[!ht]
\centering
\begin{tabular}{cccccccc}
\toprule
Digits & \multicolumn{3}{c}{Accuracy Score} &  Precision Score & Recall Score & F1 Score & AUPRC  \\
& Top-1 & Top-2 & Top-3 & Weighted & Weighted & Weighted & Weighted  \\
\midrule
2 & 0.5253 & 0.6842 & 0.7841 & 0.5825 & 0.4785 & 0.4986 & 0.5078 \\
4 & 0.3099 & 0.4250 & 0.5038 & 0.3302 & 0.2643 & 0.2522 & 0.2047 \\
6 & 0.2074 & 0.2951 & 0.3647 & 0.1930 & 0.1644 & 0.1524 & 0.1913 \\
\bottomrule
\end{tabular}
\caption{Performance of Unsupervised Industry Classification}
\label{tab:unsup-result}
\end{table}

As one of the comparable benchmark, BEACON reports a Top-1 accuracy of 0.415 and a Top-3 accuracy of 0.614 for 6-digit NAICS code classification (\textit{Ensemble}, Table 4 in \citet{dumbacherBEACONToolIndustry2025}). In contrast, our unsupervised approach exhibits a noticeable performance gap relative to BEACON. It is important to note, however, that BEACON relies on a substantially larger training dataset (i.e., millions of companies), which provides sufficient and reliable data to support the construction of large ensemble models through supervised learning. Despite this gap, unsupervised methods remain valuable in practical settings where labeled data and computational resources are limited. Furthermore, motivated by the recent surge of LLMs as the first-choice NLP solution across various domains, we also investigate their potential for industry classification and consider it as another comparable benchmark. A comparison of its performance against our unsupervised learning approach is presented in Table \ref{tab:unsup-comp}. Our experiments reveal that general-purpose LLMs, without task-specific fine-tuning or modification, struggle even to correctly interpret and execute the industry classification task. In contrast, our unsupervised learning solution remains the more reliable choice: it is substantially more computationally efficient and provides a practical and interpretable decision-support system with measurable prediction probabilities. For a detailed description of the locally deployed open-source LLM experiments using prompt engineering and Retrieval-Augmented Generation (RAG) enhanced LLM, please refer to Appendix \ref{appendix_sec:llm}. Due to data privacy, we opt out of using API calls from various commercial LLMs.
Furthermore, as the original dataset is proprietary, we construct a synthetic dataset using an LLM to facilitate reproducibility and to evaluate the proposed unsupervised learning framework. Additional details on the synthetic study are provided in Appendix \ref{appendix_sec:syn}.

\begin{table}[!ht]
\centering
\begin{tabular}{ccccc}
\toprule
Digits & Modeling & \multicolumn{3}{c}{Accuracy Score} \\
& & Top-1 & Top-2 & Top-3   \\
\midrule
\multirow{3}*{2} & Unsupervised (our) & \textbf{0.5253} & \textbf{0.6842} & \textbf{0.7841}\\
& LLM & 0.2306 & 0.2766 & 0.3027 \\
& LLM + RAG & 0.1716 & 0.2121 & 0.2243 \\
\midrule
\multirow{3}*{4} & Unsupervised (our) & \textbf{0.3099} & \textbf{0.4250} & \textbf{0.5038} \\
& LLM & 0.0767 & 0.1050 & 0.1218 \\
& LLM + RAG & 0.2970 & 0.3693 & 0.3988 \\
\midrule
\multirow{3}*{6} & Unsupervised (our) & 0.2074 & 0.2951 & \textbf{0.3647} \\
& LLM & 0.0346 & 0.0510 & 0.0637 \\
& LLM + RAG & \textbf{0.2470} & \textbf{0.3090} & 0.3419 \\
\bottomrule
\end{tabular}
\caption{Performance Comparison of Unsupervised and LLM Approaches}
\label{tab:unsup-comp}
\end{table}

Through unsupervised industry classification, we are able to objectively assign a relative industry label to a business based on the underlying textual data. This approach can serve as the foundation for an automated classification tool or as a decision-support system to assist underwriters in determining the appropriate industry classification. By minimizing human bias and inconsistency, such a system enhances both efficiency and accuracy in underwriting workflows.

Moreover, we observe that institutions such as the U.S. Census Bureau and the Internal Revenue Service have shown interest in automated industry classification, recognizing its potential value for regulatory and economic analysis. This has led to a growing body of literature in fields such as economics, public policy, and computer science. These interdisciplinary efforts highlight the broader relevance and applicability of industry classification systems and underscore the importance of continued research and development in this area.

\section{Conclusion and Discussion} \label{subsec:conclude}

In this study, we have examined the evolving role of NLP in insurance analytics, particularly through the lens of InsurTech and its alternative data sources. As insurers increasingly seek to enhance their predictive capabilities and operational efficiency, the ability to extract meaningful information from unstructured text has become vital. To systematically introduce NLP to the actuarial and insurance community, we approach NLP from a mathematical perspective that integrates various methodologies into a unified narrative, highlighting its foundations in probabilistic modeling or extensions thereof. This foundation enables the conversion of unstructured text into structured insights that can directly inform actuarial and business decision-making.

Our analysis emphasized that NLP is more than a technical add-on; it is a transformative enabler of modern insurance practices. By integrating advanced techniques such as sentiment analysis, named entity recognition, and topic modeling, insurers can deepen their understanding of risk exposure, uncover latent risk factors, and refine assessments of existing features. These techniques enable the enrichment of internal datasets using alternative sources like social media, business websites, and online reviews—unlocking new dimensions of risk profiling and customer insight.
The actuarial literature has extensively shown that using NLP to extract meaningful risk factors from unstructured text can enhance downstream actuarial analytics. In this paper, we further highlight an overlooked aspect of feature engineering: NLP can help de-bias existing numerical features by supplementing them with context-aware textual insights, and it can transform high-cardinality categorical variables into dense vector representations, thereby mitigating the curse of dimensionality. In addition to feature engineering, we also illustrate potential standalone NLP applications, to set a sprat to catch a mackerel, classifying industries for business based on textual descriptions. Collectively, these approaches enable a more data-informed strategy across core insurance functions, including pricing, underwriting, and claims handling. 
Validated by positive feedback from our InsurTech partner and its subsequent adoption by insurers, our approach demonstrates the high utility of translational research. It fosters a critical synergy between academia and industry, encouraging the integration of rigorous academic research into commercial insurance workflows.
In addition, our exploration highlights NLP's paradigm-shifting potential as a core analytical tool across a wide range of insurance applications, including automated underwriting, fraud detection, customer sentiment analysis, and litigation risk prediction, to name a few. These use cases underscore how NLP can bring both innovation and operational efficiency to the insurance industry.

Despite its potential, the adoption of NLP in actuarial science remains in its early stages. Much of the existing actuarial literature has yet to fully incorporate NLP as a foundational methodology, even as its applications expand rapidly in the industry. There is significant room for future research to explore standalone NLP use cases such as real-time claim triage, automated compliance review, personalized policy design, and behavioral risk assessment derived from communication history and textual claims data.

Due to limitations in data availability and space constraints, this paper has focused on selected use cases that we jointly explored with our InsurTech partner. Nevertheless, the examples presented underscore NLP's value as a core analytical tool for the insurance sector. We encourage researchers and practitioners to build upon this work and explore the broader transformative potential of NLP in shaping the future of insurance—driving smarter decision-making, more equitable outcomes, and greater operational agility in an increasingly data-rich world.

\newpage

\section*{Acknowledgment} The experiments on LLM industry classification used the DeltaAI advanced computing and data resource, which is supported by the National Science Foundation (award OAC 2320345) and the State of Illinois. DeltaAI is a joint effort of the University of Illinois Urbana-Champaign and its National Center for Supercomputing Applications.

\printbibliography

\newpage
\appendix

\section{Mathematical Notations}
\label{appendix_sec:notation}
\begin{longtable}{@{}ll@{}}
\toprule
Symbol & Description \\
\midrule
\endfirsthead

\multicolumn{2}{r}{{Continued from previous page}} \\
\toprule
Symbol & Description \\
\midrule
\endhead

\midrule
\multicolumn{2}{r}{{Continued on next page}} \\
\endfoot

\bottomrule
\\
\caption{Summary of symbols and their descriptions} \label{tab:notation} \\
\endlastfoot

$D$ & The corpus, set of all the documents. \\
$d$ & The document in a corpus. \\
$t$ & The term in a document. \\
$N_{td}$ & Number of times term $t$ appears in a document $d$. \\
$N_d$ & Total number of terms in a document $d$. \\
$n$ & Number of documents in a corpus $D$. \\
$m$ & Number of selected terms to represent document $d$. \\
$N_t$ & Number of documents in corpus that contain term $t$. \\
$TF-IDF$ & Term Frequency Inverse Document Frequency. \\
$TF$ & Term Frequency (in TF-IDF). \\
$IDF$ & Inverse Document Frequency (in TF-IDF). \\
\multirow{3}{*}{$A_D$} & Term-document matrix that contains the frequency of terms in documents within \\
& a corpus $D$ with $m$ terms and $n$ documents, which has components $a_{ij}$ being the \\
& frequency of term $i$ in document $j$. \\

$U$ & Matrix with columns represent the term vectors. \\
$V$ & Matrix with columns represent the document vectors. \\
$\Sigma$ & Diagonal matrix with singular values. \\

$A_k$ & Lower-dimensional matrix approximation for $A$. \\
$U_k$ & Lower-dimensional matrix approximation for $U$. \\
$V_k$ & Lower-dimensional matrix approximation for $V$. \\
$\Sigma_k$ & Diagonal matrix with $k$ largest singular values. \\

$C$ & Number of neighboring contexts for the specific term. \\
$v_{t_i}$ & The vector representation for the term $t_i$. \\

$f$ & The function notation for the algorithm. \\
$\textbf{E}$ & The embedding for the input terms.  \\
$\textbf{W}$ & The weight matrix in NN. \\
$\textbf{b}$ & The bias term in NN. \\

$z$ & The topic in a document. \\
$Z$ & Number of topics in a document. \\
$\theta$ & The topic distribution for the document. \\
$\phi$ & The word distribution for the topic. \\
$\mathcal{L}$ & A lexicon dictionary. \\
$\lambda$ & Number of top similar pairs between embedding between business and industry. \\

$y$ & The response vector in a dataset. \\
$\hat{y}$ & The model predictions. \\
$y_{[n]}$ & The sorted true response variable. \\
$J$ & Size of the dataset. \\
$\hat{y}^{Q}$ & All top-$Q$ model predictions. \\
$R$ & Number of unique categories in the multi-class classification problems. \\

$SR$ & Business star ratings which is an integer number from 1 to 5. \\
$SC$ & Sentiment score for a document. \\
$RC$ & Risk characteristics for a business. \\
$PS$ & Polarity score for a document. \\

\end{longtable}

\clearpage

\section{List of Text Cleaning Techniques}\label{appendix_sec:cleaning}

The common text cleaning techniques include:

\begin{itemize}
    \item \textbf{Contractions}: Eliminating contractions is crucial as they often involve punctuation. Failure to expand contractions may result in typo-like words. For example, \emph{I'm} becoming \emph{Im}, a non-existent English word. To deal with contractions, a dictionary of common English contractions can be employed to map contractions to their expanded forms. Besides, insurance practitioners commonly use industry-specific jargon, necessitating the manual creation of a mapping tailored to the insurance domain.
    \item \textbf{Letter case}: Normalize all words to lowercase for standardized processing. This ensures uniform treatment of words, preventing the model from distinguishing between uppercase and lowercase versions of the same word.
    \item \textbf{Filtering}: Removing words that may be irrelevant or meaningless based on various tasks is a common practice. Standard procedures include removing punctuation, nonsense symbols, and digits. For example, some InsurTech textual data originates from web scraping and contains HTML formatting syntax. These elements generally do not provide meaningful information for NLP tasks and should be removed through filtering techniques. This procedure can generally be accomplished with a few lines of \textit{regular expressions}\footnote{A method utilizes string-searching algorithms to identify sequences of characters that match a specified pattern in the text.} parsing through the corpus. 
    \item \textbf{Stopwords}: Depending on specific scenarios, it may be desirable to remove \textit{stopwords}, which are words that typically contribute minimal or no substantive content to the context, like articles, prepositions, etc. The stopwords can be derived from a general dictionary, such as the NLTK\footnote{https://www.nltk.org/} library stopwords list, or a self-defined dictionary based on the insurance domain knowledge. In the insurance context, certain frequently occurring words may lack contextual significance, while rarely appearing words could also be statistically irrelevant, so stopword removal should be conducted case by case.
\end{itemize}   

\newpage

\section{Neural LM-based Sentiment Analysis}\label{appnedix_sec:sentiment}

\begin{figure}[!ht]
\centering
\hfill
\subfigure[California vs. Florida]{\includegraphics[scale=0.75]{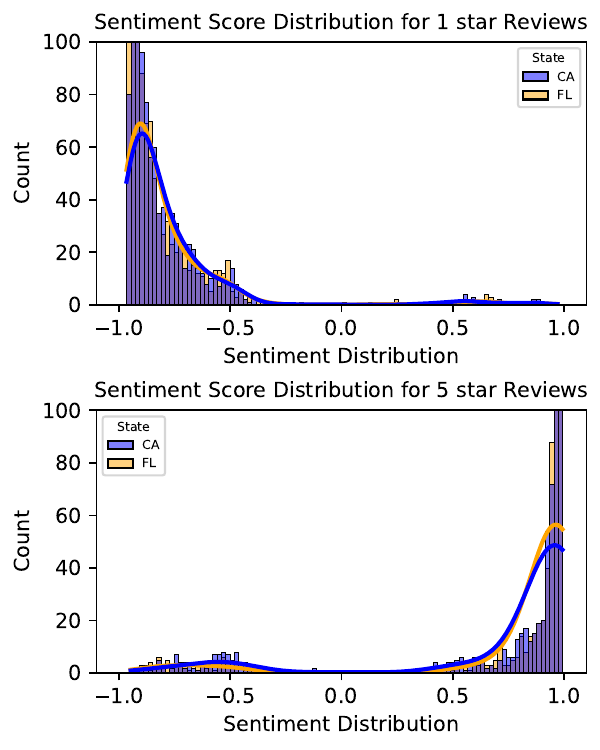}}
\centering
\hfill
\subfigure[California vs. New Jersey]{\includegraphics[scale=0.75]{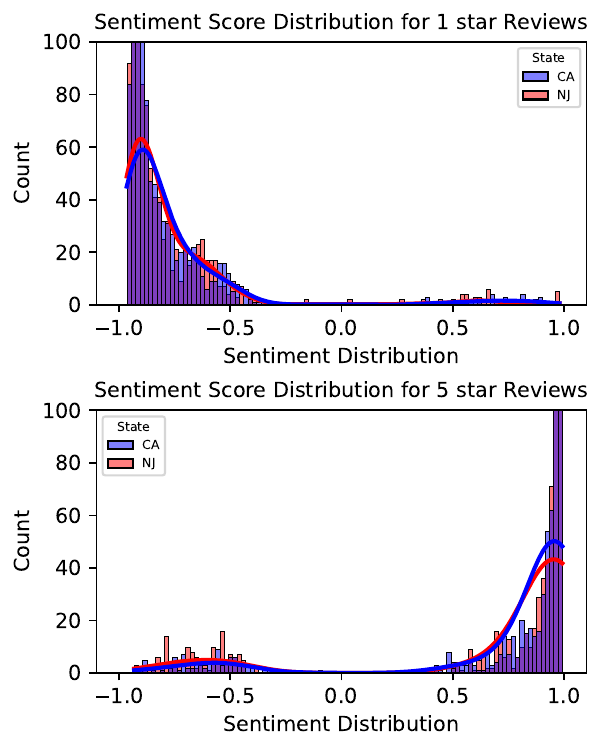}}
\hfill
\caption{Comparison of Review Sentiment Scores by Star Ratings: California vs. Other States}
\label{fig:pol-state-lm}
\end{figure}

To evaluate neural LM–based sentiment analysis, we compute sentiment classifications (i.e., positive or negative) along with their associated confidence scores (ranging from 0 to 1) using several widely adopted models. We then assign a positive or negative sign to the confidence score accordingly to construct a scalar sentiment score for this analysis. It is important to note that these scores do not directly capture sentiment intensity; rather, they reflect the model’s confidence in its classification decision. Across the evaluated models, we observe consistently high confidence levels and similar predictive behavior. We therefore report results only from \texttt{twitter-roberta-base-sentiment-latest} \citep{camacho-collados-etal-2022-tweetnlp}. Figure \ref{fig:pol-state-lm} presents the resulting sentiment scores in the state-level comparison discussed in Section \ref{subsubsec:senti}.

We observe that the sentiment scores derived from neural LMs are substantially more polarized than the lexicon-based polarity scores. In Figure \ref{fig:pol-state}, the polarity scores for 5-star reviews are centered around 0.3, while those for 1-star reviews are centered near 0. In contrast, Figure \ref{fig:pol-state-lm} shows that the neural model-based scores are concentrated near 1 and -1 for 5-star and 1-star reviews, respectively, indicating highly confident classification of positive and negative sentiment. To further examine distributional differences, we conduct K–S tests on the sentiment scores. The p-values for 5-star reviews comparing California with Florida and New Jersey are 0.0615 and 0.0005, respectively, which are substantially larger than those obtained from the lexicon-based polarity score tests (Table \ref{tab:ks}). These results suggest that, although the neural models exhibit high confidence in sentiment classification, they are less sensitive to subtle semantic variations in the reviews and may fail to capture the geographic bias observed in the data.

\newpage

\section{Evaluation Metrics}
\label{appendix_sec:eval}

\begin{enumerate}
\item Gini index:
$$
Gini(y, \hat{y})= 1 - \dfrac{2}{J-1} \left(J - \dfrac{\sum_{j=1}^{J} jy_{[j]}}{\sum_{j=1}^{J} y_{[j]}}\right)
$$

\item Percentage Error (PE)
$$
PE(y, \hat{y})=\dfrac{\sum_{j=1}^{J}(y_{j}-\hat{y}_{j})}{\sum_{j=1}^{J}y_{j}}
$$

\item Mean Absolute Error (MAE)
$$
MAE(y, \hat{y})=\dfrac{\sum_{j=1}^{J}|y_{j}-\hat{y}_{j}|}{J}
$$

\item Root Mean Squared Error (RMSE)
$$
RMSE(y, \hat{y})=\sqrt{\dfrac{\sum_{j=1}^{J}(y_{j}-\hat{y}_{j})^{2}}{J}}
$$

\item Top-$Q$ Accuracy Score
$$
Acc(y, \hat{y}^{Q})=\dfrac{\sum_{j=1}^{J}\sum_{q=1}^{Q}\mathbbm{1}_{\{y_{j}=\hat{y}^{q}_{j}\}}}{J}
$$

\item Precision Score
$$
Pr_{\text{weighted}}(y, \hat{y})=\dfrac{\sum_{r=1}^{R}J_{r}Pr_{r}}{J}
$$
where $J_{r}$ denotes the number of observations in category $r$ ($J=\sum_{r=1}^{R}J_{r}$), $Pr_{r}=\dfrac{TP_{r}}{TP_{r}+FP_{r}}$ is the precision score of category $r$. For category $r$, true positive (TP) and false positive (FP) can be written as $TP_{r}=\sum_{j=1}^{J}\mathbbm{1}_{\{y_{j}=r\}}\mathbbm{1}_{\{\hat{y}_{j}=r\}}$ and $FP_{r}=\sum_{j=1}^{J}\mathbbm{1}_{\{y_{j}\neq r\}}\mathbbm{1}_{\{\hat{y}_{j}=r\}}$ respectively.

\item Recall Score
$$
Re_{\text{weighted}}(y, \hat{y})=\dfrac{\sum_{r=1}^{R}J_{r}Re_{r}}{J}
$$
where $Re_{r}=\dfrac{TP_{r}}{TP_{r}+FN_{r}}$ is the recall score of category $r$. For category $r$, false negative (FN) can be expressed as
$$
FN_{r}=\sum_{j=1}^{J}\mathbbm{1}_{\{y_{j}\neq r\}}\mathbbm{1}_{\{\hat{y}_{j}\neq r\}}
$$

\item F1 Score
$$
F1_{\text{weighted}}(y, \hat{y})=\dfrac{\sum_{r=1}^{R}J_{r}F1_{r}}{J}
$$
where for category $r$, $F1_{r}=2\dfrac{Pr_{r}Re_{r}}{Pr_{r}+Re_{r}}$

\item Area Under Precision-Recall Curve (AUPRC)
$$
AUPRC_{\text{weighted}}(y, \hat{y})=\dfrac{\sum_{r=1}^{R}J_{r}AUPRC_{r}}{J}
$$
where for category $r$, $AUPRC_{r}=\sum_{i}(Re_{r}^{i}-Re_{r}^{i-1})Pr_{r}^{i}$ with $i$-th thresholds of recall and precision $Re_{r}^{i}$ and $Pr_{r}^{i}$, respectively.
\end{enumerate}

\clearpage

\section{Setting of LLM Industry Classification}\label{appendix_sec:llm}

The rapid rise of LLM-powered Generative AI (GenAI) systems has enabled a wide range of applications, particularly in text generation. A natural question is whether LLMs can also assist in industry classification and how their performance compares to our unsupervised learning solution. To investigate this, we design a series of experiments on LLM-based industry classification. Due to the sensitivity of our dataset, we restrict our evaluation to open-source models rather than calling commercial closed-source APIs. In particular, we adopt the 32B variant of DeepSeek-R1 \citep{guoDeepSeekR1IncentivizesReasoning2025}, one of the strongest performing open-source LLMs, to generate classifications based on firm categorizations and descriptions.

To enable more systematic evaluation, we prompt the model to output the top-3 most probable NAICS codes rather than textual industry labels, with the exact prompt illustrated in Figure \ref{fig:prompt}. Despite this instruction, the model occasionally fails to adhere to the required output format, for instance, producing six-digit codes when only two-digit predictions are requested, or returning textual labels instead of numerical codes. In such cases, we substitute placeholders (e.g., ``00" for two-digit outputs) to explicitly mark incorrect predictions. We also observe cases where the model generates non-existent NAICS codes (e.g., 17, 59, 70). Further inspection of the model outputs indicates that the model often becomes trapped in spurious reasoning about code ordering or digit formatting, as illustrated in Figure \ref{fig:llm-exam}.

\begin{figure}[!ht]
\centering
\fbox{%
\begin{minipage}{0.8\textwidth}
\textbf{role}: system\\
\textbf{content}: You are a helpful assistant for insurance companies. Please give short and straightforward answers.\\\\
\textbf{role}: user \\
\textbf{content}:\\
Context: [RAG CONTENT IF RAG IS ENABLED]\\
Please classify the business using the North American Industry Classification System (NAICS) based on the following content and category information provided by the business. Provide the top-3 most probable [DIGIT]-digit codes. Use the context above if it is helpful.
\\\\
Business content: [CONTENT]\\
Business category: [CATEGORY]
\\\\
Please parse ``answer" and output them in JSON format.
\\\\
EXAMPLE JSON OUTPUT:\\
\{\{\\
    ``First NAICS code": ``CODE 1",\\
    ``Second NAICS code": ``CODE 2",\\
    ``Third NAICS code": ``CODE 3"\\
\}\}
\end{minipage}
}
\caption{Design of Industry Classification Prompt}\label{fig:prompt}
\end{figure}

\begin{figure}[!ht]
\centering
\fbox{%
\begin{minipage}{0.8\textwidth}
Alternatively, perhaps 51, 54, and 51 again, but I think the user wants distinct codes. So, I'll go with 51 as the first, 54 as the second, and maybe 52 if they have any financial activities, but I don't see that. Alternatively, perhaps 51, 54, and 51 again, but that's not helpful.\\
...\\
Wait, perhaps the third code is 31, but I'm not certain. Alternatively, maybe 31 is not necessary, and the third code could be another code related to chemicals, but I don't think there's a 3-digit code that would be a 2-digit code.\\
Alternatively, perhaps the third code is 31, but I'm not certain. Alternatively, maybe 31 is not necessary, and the third code could be another code related to chemicals, but I don't think there's a 3-digit code that would be a 2-digit code.\\
...\\
Alternatively, perhaps the third code is 31, but I'm not certain. Alternatively, maybe 31 is not necessary, and the third code could be another code related to chemicals, but I don't think there's a 3-digit code that would be a 2-digit code.
\end{minipage}
}
\caption{Examples of LLM Stuck in Reasoning}\label{fig:llm-exam}
\end{figure}

To mitigate these issues, we incorporate Retrieval-Augmented Generation (RAG) with the official NAICS documentation. In this setup, RAG retrieves relevant sections of the NAICS codebook based on embedding similarity and injects the extracted text into the LLM prompt, thereby grounding the model with precise reference material. The results of vanilla LLM and RAG-enhanced LLM industry classification are summarized in Table \ref{tab:unsup-comp}. The results indicate that LLMs themselves struggle to deliver accurate industry classification. Incorporating retrieved information substantially improves performance on 4- and 6-digit classification tasks, but has a mixed effect on 2-digit classification. Specifically, retrieved documents, being more detailed and business-specific, tend to encourage the model to output 6-digit codes even when only 2-digit predictions are required, thereby reducing accuracy in that setting. In contrast, their contribution is much more beneficial for the finer-grained 4- and 6-digit tasks. Another notable observation is that LLM-based classification relies much more heavily on the top-1 prediction. Whereas our unsupervised learning solution exhibits a roughly 20\% gap between top-1 and top-3 accuracy, LLM-based performance shows less than a 10\% gap, highlighting their limited ability to provide reliable alternatives beyond the single most probable output.

Overall, our experiments demonstrate that vanilla LLMs substantially underperform compared to our unsupervised learning solution, whereas RAG-enhanced LLMs achieve competitive results and, in certain granular tasks, even outperform the unsupervised approach. In addition, our unsupervised approach is far more computationally efficient: it completes in approximately one CPU hour for all digit-level tasks, whereas LLM-based classification requires roughly 24 GPU hours on an H200 for each digit-level experiment (excluding text preprocessing in both cases). These findings suggest that, although LLMs are widely considered state-of-the-art, they remain less effective in applications requiring structured domain knowledge and precise decision-making, unless explicitly modified or fine-tuned for the target task, which requires substantial training data.

\newpage

\section{Synthetic Business Industry Classification}\label{appendix_sec:syn}

To facilitate reproducibility and provide a clearer understanding of the unsupervised business industry classification framework described in Section \ref{subsubsec:class}, we construct a synthetic business information dataset. Given the textual nature of the data, we employ an LLM to generate the required business information. As illustrated in Figure \ref{fig:unsup-pipe}, the dataset integrates information from two sources: NAICS and business-level data. The NAICS information, including the 6-digit code, title, and description, is publicly available and extracted from the U.S. Census Bureau. For each of the 2,372 businesses in our dataset, we have access to the confirmed NAICS code, business description (\textit{content}), and InsurTech categories. To generate the synthetic data, we prompt the LLM using the business’s 6-digit NAICS code, along with its associated title and description, and instruct the model to produce a synthetic business description and a set of relevant category labels. It is important to note that this prompt-based generation process conditions on the ground-truth NAICS information, which expose the response variable and thereby improve downstream classification performance. In contrast, real-world business data are often noisy, unstructured, and contain redundant or irrelevant information, making the classification task substantially more challenging. The prompt to generate the synthetic business information is illustrated in Figure \ref{fig:prompt-data-gen}.

\begin{figure}[!ht]
\centering
\fbox{%
\begin{minipage}{0.8\textwidth}
\textbf{role}: system\\
\textbf{content}: You are a helpful assistant to generate synthetic data for insurance industry classification.\\\\
\textbf{role}: user \\
\textbf{content}:\\
Please generate synthetic business information of business content and categories using the North American Industry Classification System (NAICS) based on the following NAICS information provided by the business. \\\\
The business content is a long paragraph of sentences that describe the business's scope/coverage.\\
The business categories are a short list of words that the business used to categorize themselves.\\\\
The business's naics code: {naics}\\
The naics's title: {title}\\
The naics's description: {description}\\\\
Please parse "answer" and output them in JSON format.\\\\
EXAMPLE JSON OUTPUT:\\
\{\{\\
    "content": [CONTENT],\\
    "categories": [CATEGORIES],\\
\}\}\\
\end{minipage}
}
\caption{Design of Synthetic Data Generation Prompt}\label{fig:prompt-data-gen}
\end{figure}

As summarized in Table \ref{tab:unsup-syn}, the unsupervised learning framework achieves high accuracy in identifying NAICS categories, largely due to the nature of the data generation process. For example, the 6-digit NAICS code 311920 corresponds to \textit{coffee and tea manufacturing} in the official documentation, and the synthetic business description explicitly includes phrases such as “manufacturing of coffee and tea products,” along with category labels such as “Coffee Manufacturing” and “Tea Manufacturing.” This direct alignment between textual descriptions and category labels facilitates accurate classification. Furthermore, such explicit category encoding disproportionately improves the 6-digit classification accuracy as the synthetic businesses are generated from the 6-digit NAICS information. However, such clear and structured relationships are rarely observed in real-world applications. Consequently, as shown in Table \ref{tab:unsup-comp}, classification performance is substantially lower on real data. These results suggest that the proposed unsupervised framework relies heavily on embedding similarity, and that data quality plays a critical role in its effectiveness in practical settings.

\begin{table}[!ht]
\centering
\begin{tabular}{cccc}
\toprule
Digits & \multicolumn{3}{c}{Accuracy Score} \\
& Top-1 & Top-2 & Top-3   \\
\midrule
2 & 0.7879 & 0.8841 & 0.9165\\
4 & 0.7449 & 0.8596 & 0.8976 \\
6 & 0.8137 & 0.8971 & 0.9228 \\
\bottomrule
\end{tabular}
\caption{Performance of Unsupervised Learning on Synthetic Data}
\label{tab:unsup-syn}
\end{table}

\end{document}